\documentclass{article}

\usepackage[final,nonatbib]{neurips_2022}

\usepackage[utf8]{inputenc} %
\usepackage[T1]{fontenc}    %
\usepackage{graphicx}
\usepackage{tikz}
\usepackage{url}            %
\usepackage{booktabs}       %
\usepackage{amsfonts}       %
\usepackage{amsmath}
\usepackage[binary-units]{siunitx}
\usepackage{wrapfig}
\usepackage{subcaption}
\usepackage{nicefrac}       %
\usepackage{microtype}      %
\usepackage{xcolor}         %
\usepackage{xspace}
\usepackage{enumitem}
\usepackage{hyperref}       %

\usepackage{soul}

\usepackage{biblatex} %
\newcommand{\name}{VCT}

\newcommand*{\zjoint}{z_\text{joint}}
\newcommand*{\zcur}{z_\text{cur}}

\newcommand*{\eg}{\textit{e.g.}}
\newcommand*{\ie}{\textit{i.e.}}
\newcommand*{\etal}{\textit{et al.}\xspace}

\newcommand*{\cur}{\text{c}}
\newcommand*{\prev}{\text{p}}

\newcommand*{\numc}{d_C}
\newcommand*{\dmodel}{d_T}

\newcommand*{\Tsep}{T_\text{sep}}
\newcommand*{\Tjoint}{T_\text{joint}}
\newcommand*{\Tcur}{T_\text{cur}}

\definecolor{ibmpurple}{rgb}{0.47,0.37,0.94}
\definecolor{ibmpink}{rgb}{0.86,0.15,0.50}

\newcolumntype{L}[1]{>{\raggedright\let\newline\\\arraybackslash\hspace{0pt}}m{#1}}
\newcolumntype{C}[1]{>{\centering\let\newline\\\arraybackslash\hspace{0pt}}m{#1}}
\newcolumntype{R}[1]{>{\raggedleft\let\newline\\\arraybackslash\hspace{0pt}}m{#1}}

\newcommand{\smemail}[1]{\footnotesize\texttt{#1}}

\title{VCT: A Video Compression Transformer}

\author{%
  Fabian Mentzer \\
  Google Research \\
  \smemail{mentzer@google.com} \\
  \And
  George Toderici \\
  Google Research \\
  \smemail{gtoderici@google.com} \\
  \And
  David Minnen \\
  Google Research\\
  \smemail{dminnen@google.com} \\
  \And
  Sung Jin Hwang \\
  Google Research \\
  \smemail{sjhwang@google.com} \\
  \And
  Sergi Caelles \\
  Google Research \\
  \smemail{scaelles@google.com} \\
  \And
  Mario Lucic \\
  Google Research \\
  \smemail{lucic@google.com} \\
  \And
  Eirikur Agustsson \\
  Google Research \\
  \smemail{eirikur@google.com} \\
  }

\makeatletter
\renewcommand{\paragraph}{%
  \@startsection{paragraph}{4}%
  {\z@}{0.25ex \@plus 0.1ex \@minus 0.2ex}{-1em}%
  {\normalfont\normalsize\bfseries}%
}
\makeatother

\begin{document}

\maketitle

\begin{abstract}
We show how transformers can be used to vastly simplify neural video compression. Previous methods have been relying on an increasing number of architectural biases and priors, including motion prediction and warping operations, resulting in complex models. Instead, we independently map input frames to representations and use a transformer to model their dependencies, letting it predict the distribution of future representations given the past. The resulting video compression transformer outperforms previous methods on standard video compression data sets. Experiments on synthetic data show that our model learns to handle complex motion patterns such as panning, blurring and fading purely from data. Our approach is easy to implement, and we release code to facilitate future research.
\end{abstract}

\section{Introduction}

Neural network based video compression techniques have recently emerged to rival their non-neural counter parts in rate-distortion performance~\cite[\eg,][]{agustsson2020scale,hu2021fvc,rippel2021elf,yang2020hierarchical}.
These novel methods tend to incorporate various architectural biases and priors inspired by the classic, non-neural approaches.
While many authors tend to draw a line between ``hand-crafted'' classical codecs and neural approaches, the neural approaches themselves are increasingly ``hand-crafted'', with authors introducing complex connections between the many sub-components.
The resulting methods are complicated, challenging to implement,
and constrain themselves to work well only on data that matches the architectural biases.
In particular, many methods rely on some form of motion prediction followed by a warping operation~\cite[\eg,][]{agustsson2020scale,hu2021fvc,li2021deep,lu2019dvc,yang2020hierarchical,yang2021rlvc}. 
These methods warp previous reconstructions with the predicted flow, and calculate a residual.

In this paper, we replace flow prediction, warping, and residual compensation, with an elegantly simple but powerful transformer-based temporal entropy model.
The resulting video compression transformer (VCT) outperforms previous methods on standard video compression data sets, while being free from their architectural biases and priors.
Furthermore, we create synthetic data to explore the effect of architectural biases, and
show that we compare favourably to previous approaches on the types videos that the architectural components were designed for (panning on static frames, or blurring), despite our transformer not relying on any of these components. 
More crucially, we outperform previous approaches on videos that have no obvious matching architectural component (sharpening, fading between scenes), showing the benefit of removing hand-crafted elements and letting a transformer learn everything from data.

\begin{figure}
	\centering
	\includegraphics[width=0.7\textwidth]{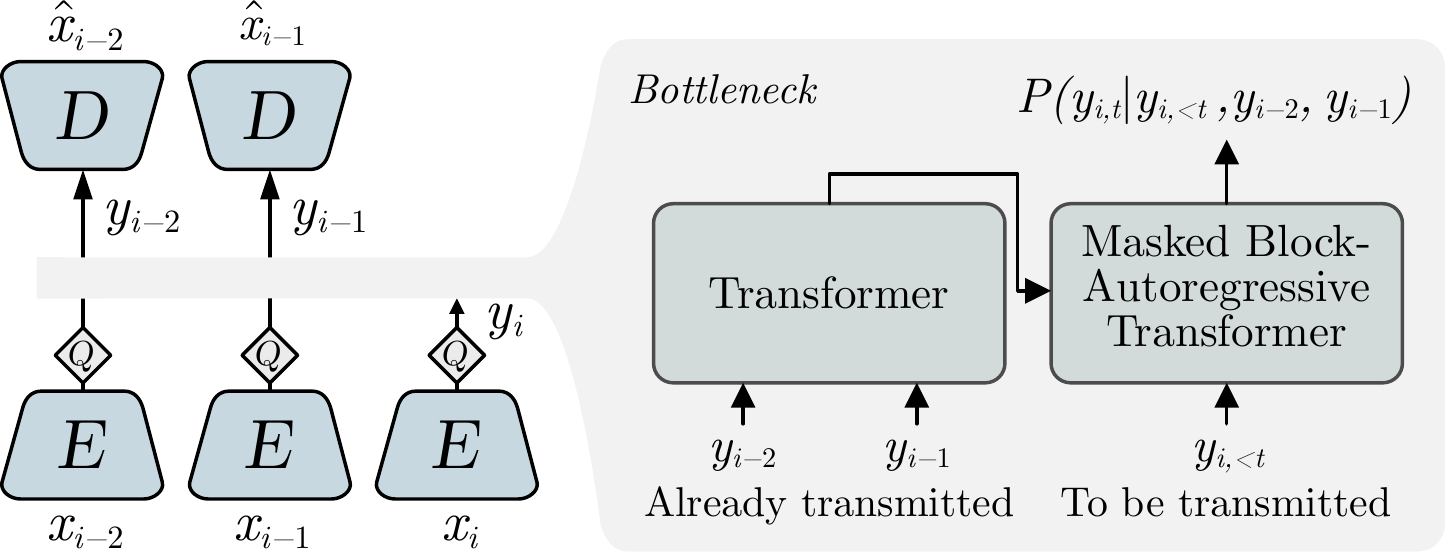}
	\caption{\label{fig:overview} We independently and \emph{lossily} map input frames $x$ into quantized representations $y$. From $y$ we can recover a reconstruction $\hat x$. 
	To store $y_i$ with few bits, we 
	use transformers to model temporal dependencies and to predict a distribution for $y_i$ given previously transmitted representations.
	We use $P$ to \emph{losslessly} compress the quantized $y_i$ using entropy coding. 
	The better the transformer predicts $P$, the fewer bits are required to store $y_i$.
	We note that we have no hard-coded components such as motion prediction or warping. }
\end{figure}

We use transformers to compress videos in two steps (see Fig.~\ref{fig:overview}): 
First, using \emph{lossy} transform coding~\cite{balle2020nonlinear}, 
we map frames $x_i$ from image space to quantized representations $y_i$, \emph{independently for each frame}. 
From $y_i$ we can recover a reconstruction $\hat x_i$. 
Second, we let a transformer leverage temporal redundancies to model the distributions of the representations.
We use these predicted distributions to \emph{losslessly} compress the quantized $y_i$ using entropy coding~\cite[Sec 2.2.1]{Yang2022a}. 
The better the transformer predicts the distributions, the fewer bits are required to store the representations.

This setup avoids complex state transitions or warping operations by letting the transformer learn to leverage arbitrary relationships between frames. As a bonus, we get rid of temporal error propagation by construction since the reconstruction $\hat x_i$ does not depend on previous reconstructions.
Contrast with warping-based approaches, where $\hat x_i$ is a function of the warped $\hat x_{i-1}$ meaning that any visual errors in $\hat x_i$ will be propagated forward and require additional bits to correct with residuals.

VCT is based on the original language translation transformer~\cite{vaswani2017attention}: We can view our problem as ``translating'' two previous representations $y_{i-2}, y_{i-1}$ to $y_i$.
However, there are various challenges in the way of directly applying the NLP formulation.
Consider a 1080p video frame; using a typical neural image compression encoder~\cite{balle2018variational} that downscales by a factor 16 and has 192 output channels, a (1080, 1920, 3)-dimensional input frame is mapped to a (68, 120, 192)-dimensional feature representation leading to approximately $1.6$ million symbols. 
Naively correlating all of these symbols to all symbols in a previous representation would yield a $1.6\text{M}{\times}1.6\text{M}$-dimensional attention matrix. To address this computationally impractical problem, we introduce independence assumptions to shrink the attention matrix and enable parallel execution on subsets of the symbols.

Our model is easy to implement with contemporary machine learning frameworks, and we provide an extensive code and model release to allow future work to build on this direction.\footnote{
\url{https://goo.gle/vct-paper}
}

\section{Related Work}

Transformers were initially proposed for machine translation~\cite{vaswani2017attention}, where an encoder-decoder structure was used to obtain state-of-the-art results. 
This led to a wide range of follow-up research, and state-of-the-art natural language processing (NLP) models are still based on transformers~\cite[\eg,][]{brown2020language,devlin2018bert,chowdhery2022palm,edunov2018understanding}.
Motivated by these advancements, Dosovitski~\etal~\cite{dosovitskiy2020image} replaced CNNs with a transformer-based architecture to achieve state-of-the-art results in image classification, which in turn spurred more exploration of transformers in the computer vision community including both image tasks~\cite[\eg,][]{liu2021swin,wang2021max,zheng2021rethinking} as well as video analysis~\cite[\eg,][]{arnab2021vivit,bertasius2021space,fan2021multiscale,neimark2021video,sun2019videobert}. 

Recently, transformers were incorporated into neural \emph{image} compression models. Qian~\etal~\cite{qian2022entroformer} replaced the autoregressive hyperprior~\cite{minnen2018joint} with a self-attention stack, and Zhu~\etal~\cite{zhu2021transformer} replaced all convolutions in the standard approach~\cite{balle2018variational,minnen2020channel} with Swin Transformer~\cite{liu2021swin} blocks.

Neural \emph{video} compression remains CNN-based. After initial work used frame interpolation~\cite{wu2018video,djelouah2019neural}, Lu~\etal~\cite{lu2019dvc} followed the more traditional approach of predicting optical flow between the previous reconstruction and the input, transmitting a compressed representation of the flow, and also transmitting a residual image to correct visual errors after warping.
Many papers extended this approach, for example Agustsson~\etal~\cite{agustsson2020scale} introduced the notion of a flow predictor that also supports blurring called ``Scale Space Flow'' (SSF), which became a building block for other approaches~\cite{yang2020hierarchical,rippel2021elf}. 
Rippel~\etal~\cite{rippel2021elf} achieved state-of-the-art results by using SSF and more context to predict flow. 
RNNs and ConvLSTMs were used to build recurrent decoders~\cite{golinski2020feedback} or entropy models~\cite{yang2021rlvc}.

Some work does not rely on pixel-space flow:
Habibian~\etal~\cite{habibian2019video} used a 3D autoregressive entropy model,
FVC~\cite{hu2021fvc} predicted flow in a $2{\times}$ downscaled feature space,
and Liu~\etal~\cite{liu2021deep} used a ConvLSTM to predict representations which are transmitted using an iterative quantization scheme.
DCVC~\cite{li2021deep} estimated motion in pixel space but performed residual compensation in a feature space.
Liu~\etal~\cite{liu2020conditional} also losslessly encoded frame-level representations, but rely on CNNs for temporal modelling. Finally, recent work employed GAN losses to increase realism~\cite{mentzer2021neural,yang2021perceptual}.

\begin{figure}[t]
	\centering
	\begin{minipage}[b][3cm]{0.2\textwidth}
	\hfill Tokens\hspace{1ex}
	\vfill
	\hfill Blocks\hspace{1ex}
	\vfill
	\hfill Representations\hspace{1ex}
	\vfill
	\phantom{x}%
	\end{minipage}
	\includegraphics[height=3.75cm]{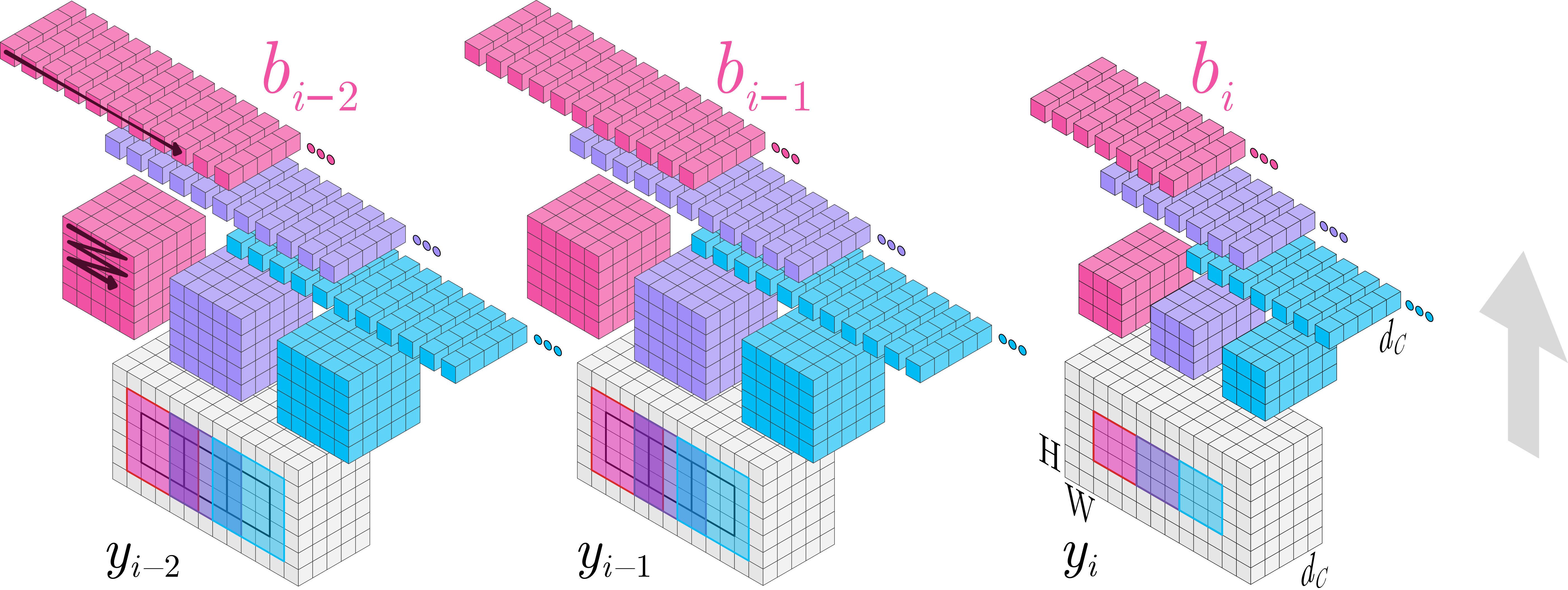}\\
	{\small
	\hspace{0.36\textwidth} \textit{ Already Transmitted \hfill To be transmitted\hspace{0.1\textwidth}}
	}
	\vspace{-.3ex}
	\caption{
\label{fig:tokens}From representations to tokens. We essentially use a sliding window to split the current representation $y_i$ into \emph{non-overlapping} $w_\cur \times w_\cur$ blocks, and previous representations $y_{i-2},y_{i-1}$ into \emph{overlapping} $w_\prev \times w_\prev$ blocks with stride $w_\cur$ ($w_\prev > w_\cur$). 
We flatten blocks spatially (raster-scan order, see left arrows) to obtain tokens for the transformer, which remain $\numc$-dimensional since they are just a different view of $y_i$.
We show $w_\cur{=}3,w_\prev{=}5,\numc{=}5$, but we use $w_\cur{=}4,w_\prev{=}8,\numc{=}192$ in practice.
}
	
\end{figure}

\section{Method}

\subsection{Overview and Background}

\paragraph{Frame encoding and decoding} A high-level overview of our approach is shown in Fig.~\ref{fig:overview}.
We split video coding into two parts. 
First, we independently encode each frame $x_i$ into a \emph{quantized} representation $y_i{=}\lfloor E(x_i) \rceil$ using a CNN-based image encoder $E$ followed by quantization to an integer grid.
The encoder downscales spatially and increases the channel dimension,
resulting in $y_i$ being a $(H,W,\numc)$-dimensional feature map, where $H,W$ are $16{\times}$ smaller than the input image resolution.
From $y_i$, we can recover a reconstruction $\hat x_i$ using the decoder $D$.
We train $E, D$ using standard neural image compression techniques to be lossy transforms reaching nearly any desired distortion $d(x_i, \hat x_i)$ by varying how large the range of each element in $y_i$ is.
For now, let us assume we have a pair $E, D$ reaching a fixed distortion.

\paragraph{Naive approach} After having \emph{lossily} converted the sequence of input frames $x_i$ to a sequence of representations $y_i{=}\lfloor E(x_i) \rceil$, one can naively store all $y_i$ to disk \emph{losslessly}.
To see why this is sub-optimal, let each element $y_{i,j}$ of $y_i$ be a symbol in $\mathcal{S}=\{-L, \dots, L\}$. Assuming that all $|\mathcal S|$ symbols appear with equal probability, \ie, $P(y_{i,j})=1/\lvert \mathcal S \rvert$, one can transmit $y_i$ using $H\cdot W\cdot \numc \cdot \log_2\lvert \mathcal S \rvert$ bits.
Using a realistic $L{=}32$, this implies that we would need \SI{9}{\mega\byte}, or $\approx$2Gbps at 30fps, to encode a single HD frame (where $H \cdot W \cdot \numc{\approx}1.6$M, see Introduction).
While arguably inefficient, this is a valid compression scheme which will result in the desired distortion. The aim of this work is to improve this scheme by approximately two orders of magnitude.

\paragraph{An efficient coding scheme} Given a probability mass function (PMF) $P$ estimating the true distribution $Q$ of symbols in $y_i$, we can use entropy coding (EC) to transmit $y_i$ with $H\cdot W\cdot \numc \cdot \mathbb{E}_{y\sim Q(y_{i})}[-\text{log}_2 P(y)]$ bits.\footnote{Consistent with neural compression literature but in contrast to Information Theory, we use $P$ for the model.}
By using EC, we can encode more frequently occurring values with fewer bits, and hence improve the efficiency. 
Note that the expectation term representing the average bit count corresponds to the cross-entropy of $Q$ with respect to $P$. 
Our main idea is to parameterize $P$ as a conditional distribution using very flexible transformer models, and to minimize the cross-entropy and thus maximize coding efficiency.
We emphasize that we use $P$ for lossless EC, we do not sample from the model to transmit data.
Even if the resulting model of $P$ is sub-optimal, $y_i$ can still be stored losslessly, albeit inefficiently. 

Why would one hope to do better than the uniform distribution over $y_i$? In principle, the model should be able to exploit  the temporal redundancy across frames, and the spatial consistency within frames.

\begin{figure}[t]
	\centering
	\includegraphics[width=\textwidth]{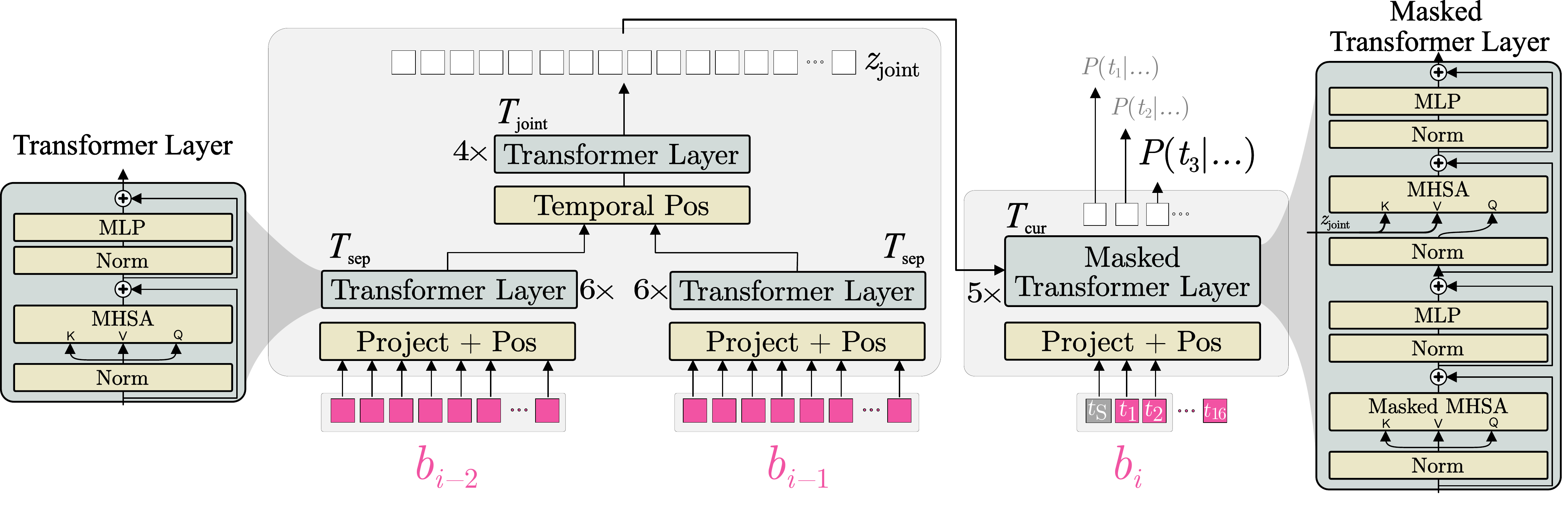} \\[-1ex]
	{\small
	\hspace{0.3\textwidth} \textit{ Already Transmitted \hfill To be transmitted\hspace{0.2\textwidth}}}
	\caption{\label{fig:details} 
	The transformer operates on the pink set of blocks/tokens $b_{i-2},b_{i-1},b_i$ (obtained as shown in Fig.~\ref{fig:tokens}). 
	We first extract temporal information $\zjoint$ from already transmitted blocks.
	$T_\text{cur}$ is shown predicting $P(t_3|t_S,t_1,t_2,\zjoint)$, where $t_S$ is a learned start token.
	}
\end{figure}

\subsection{Transformer-based Temporal Entropy Model} \label{sec:overview}
To transmit a video of $F$ frames, $x_1, \dots, x_F$, we first map $E$ over each frame obtaining quantized representations $y_1, \dots, y_F$.
Let's assume we already transmitted $y_1, \dots, y_{i-1}$. 
To transmit $y_i$, we use the transformer to predict $P(y_i|y_{i-2}, y_{i-1})$.
Using this distribution, we entropy code $y_i$ to create a compressed, binary representation that can be transmitted.
To compress the full video, we simply apply this procedure iteratively, letting the transformer predict $P(y_j|y_{j-2},y_{j-1})$ for $j\in\{1,\dots,F\}$, padding with zeros when predicting distributions for $y_1, y_2$.
The receiver follows the same procedure to recover all $y_j$, \ie, it iteratively calculates $P(y_j|y_{j-2},y_{j-1})$ to entropy decode each $y_j$. 
After obtaining each representation, $y_1, y_2, \dots, y_F$, the receiver generates reconstructions.%

\paragraph{Tokens} When processing the current representation $y_i$, we split it spatially into \emph{non-overlapping} blocks with size $w_\cur \times w_\cur$ as shown in Fig.~\ref{fig:tokens}. Previous representations $y_{i-2},y_{i-1}$ become corresponding \emph{overlapping} $w_\prev \times w_\prev$ blocks (where $w_\prev > w_\cur$) to provide both temporal and spatial context for predicting $P(y_i|y_{i-2},y_{i-1})$. Intuitively, the larger spatial extent provides useful context to predict the distribution of the current block.
Note that all blocks span a relatively large spatial region in image space due to the downscaling convolutional encoder $E$.
We flatten each block spatially (see Fig.~\ref{fig:tokens}) to obtain tokens for the transformers.
The transformers then run independently on corresponding blocks/tokens, \ie, tokens of the same color in Fig.~\ref{fig:tokens} get processed together, trading reduced spatial context for parallel execution.\footnote{%
As a side benefit, the number of tokens for the transformers is not a function of image resolution, unlike ViT-based approaches~\cite{dosovitskiy2020image}.}

This independence assumption allows us to focus on a single set of blocks, \eg, the pink blocks in Fig.~\ref{fig:tokens}.
In the following text and in Fig.~\ref{fig:details}, we thus show how we predict distributions for the $w_\cur^2{=}16$ tokens $t_1, t_2, \dots, t_{16}$ in block $b_i$,
given the $2w_\prev^2{=}128$ tokens from the previous blocks $b_{i-2},b_{i-1}$.

\paragraph{Step 1: Temporal Mixer} 
We use two transformers to extract temporal information from $b_{i-2},b_{i-1}$.
A first transformer $\Tsep$ operates separately on each previous block.
Then, we concatenate the outputs in the token dimension and run the second transformer, $\Tjoint$, on the result to mix information across time. 
The output $\zjoint$ is $2w_\prev^2$ features, containing everything the model ``knows'' about the past.

\begin{figure}[t]
	\centering
	\includegraphics[width=0.9\textwidth]{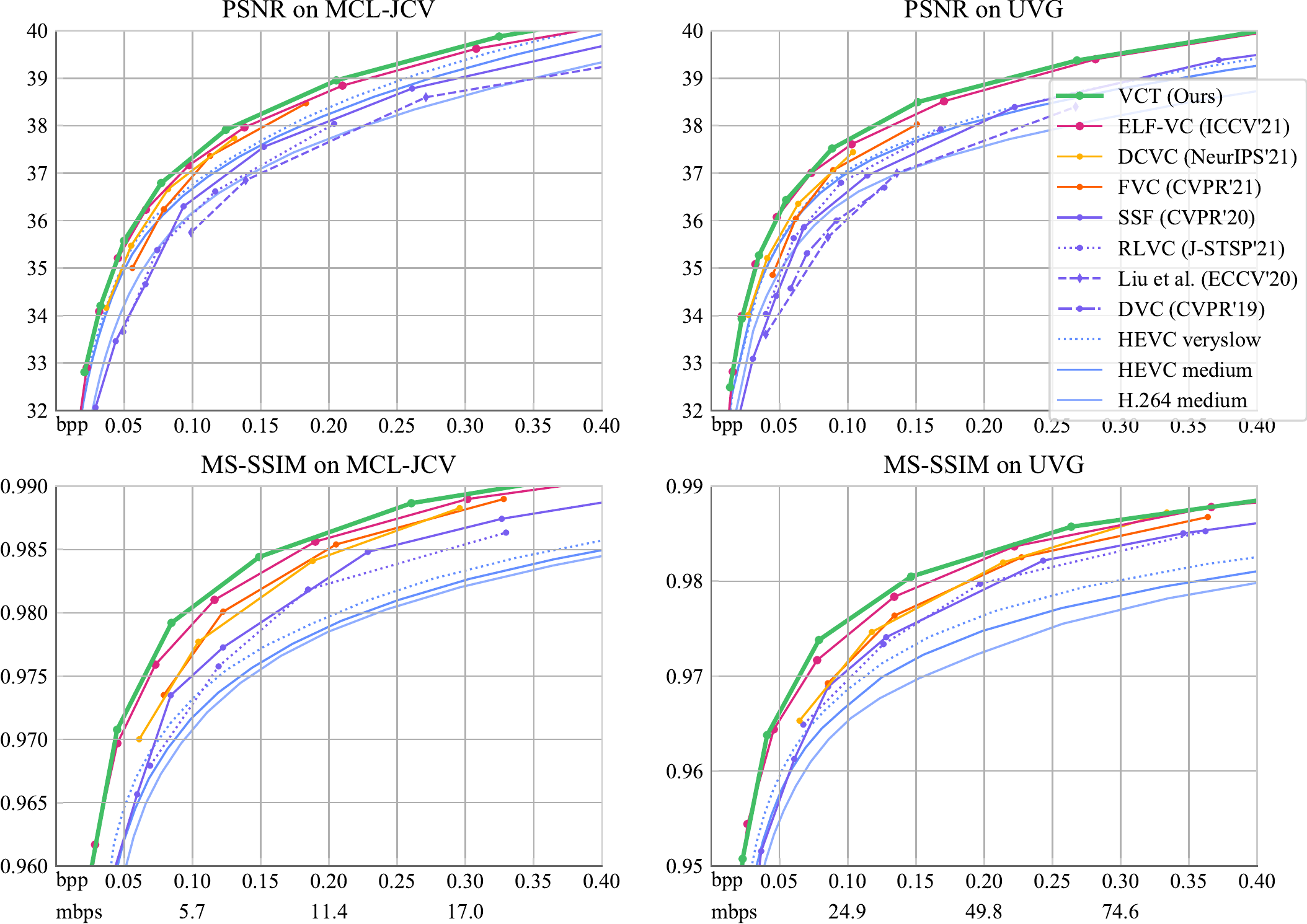}
	\caption{\label{fig:rd_results} 
	Comparing rate-distortion on MCL-JCV ($\approx$27FPS) and UVG (120FPS). We report bits per pixel (bpp) and megabits per second (mbps). For MS-SSIM, we only show methods optimized for it (using \textit{--tune ssim} for HEVC/H264). \textit{App.~\ref{sec:app:bigplots} shows a large version of these plots.}
	\vspace{-2ex}
	}
\end{figure}

\paragraph{Step 2: Within-Block-Autoregression} 
The second part of our method is the masked transformer $\Tcur$, which predicts PMFs for each token 
using auto-regression within the block.
We obtain a powerful model by conditioning $\Tcur$ on $\zjoint$ as well as already transmitted tokens within the block.
For entropy coding, both the sender and the receiver must be able to obtain exactly the same PMFs, \ie, $\Tcur$ must be causal and start from a known initialization point. For the latter, we learn a \emph{start token} $t_S$. 

To send the tokens, we first obtain $\zjoint$. After that, we feed $[t_S]$ to $\Tcur$, obtain $P(t_1|t_S ; \zjoint)$, and use entropy coding to store the $\numc$ symbols in token $t_1$ into a bitstream using $P(t_1|t_S ; \zjoint)$. Then, we feed $[t_S, t_1]$, obtain $P(t_2| t_1, t_S ; \zjoint)$, store $t_2$ in the bitstream, and so on. 
The receiver gets the resulting bitstream and can obtain the same distributions, 
and thereby the tokens, by first feeding $[t_S]$ to $\Tcur$,
obtaining $P(t_1|t_S ; \zjoint)$, entropy decoding $t_1$ from the bitstream, then feeding $[t_S,t_1]$ to obtain $P(t_2|t_1, t_S ; \zjoint)$, and so on.
Fig.~\ref{fig:details} visualizes this for $P(t_3|\dots)$.

We run this procedure in parallel over all blocks, and thereby send/receive $y_i$ by running $\Tcur$ $w_\cur^2{=}16$ times. 
Each run produces $\lceil H/w_\cur\rceil \cdot \lceil W/w_\cur \rceil \cdot \numc$ distributions.
To ensure causality of $\Tcur$ during training, we mask the self-attention blocks similar to~\cite{vaswani2017attention}.

\paragraph{Independence}
Apart from assuming blocks in $y_i$ are independent, we emphasize that each token is a vector and that we assume the symbols within each token are conditionally independent given previous tokens, \ie,
 $\Tcur$ predicts the $\numc$ distributions required for a token \emph{at once}. One could instead predict a joint distribution over all possible $|\mathcal S|^{\numc}$ realisations, use channel-autoregression~\cite{minnen2020channel}, or use vector quantization on tokens.  The latter two are interesting directions for future work.
Finally, we note that we do not rely on additional side information, in contrast to, \eg, autoregressive image compression entropy models~\cite{minnen2018joint,minnen2020channel}.

\subsection{Architectures}\label{sec:archs}

\begin{table}[b]
    \centering
    \begin{tabular}{l@{\hskip 2em}l@{\hskip 2em}lccccc}
    \toprule
          & Components trained & Loss & $B$ & $N_F$ & LR &  Steps & steps/s  \\
         \midrule
         Stage I & $E, D$ 
            & $r+\lambda d$ & 16 &  1 & $\phantom{2.}1\textsc{e}^{-4}$   & 2M   & $100$ \\
         Stage II & $\Tsep, \Tjoint, \Tcur$ 
            & $r$           & 32 &  3 & $\phantom{2.}1\textsc{e}^{-4}$   & 1M   & $\phantom{0}10$ \\
         Stage III & $\Tsep, \Tjoint, \Tcur, E, D$ 
            & $r+\lambda d$ & 32 &  3 & $2.5\textsc{e}^{-5}$ & 250k & $\phantom{00}5$ \\
         \bottomrule \\[-1ex]
    \end{tabular}
    \caption{
    \label{tab:training_setup}
    We split training in three stages for training efficiency (note the steps/s column). $\lambda$ controls the rate-distortion trade-off, $r$ is bitrate, $d$ is distortion, $B$ is batch size, $N_F$ the number of frames.
    }
\end{table}

\paragraph{Transformers} As visualized in Fig.~\ref{fig:details}, all of our transformers are based on standard architectures~\cite{vaswani2017attention,dosovitskiy2020image}.
We start by projecting the $\numc$-dimensional tokens to a $\dmodel$-dimensional space ($\dmodel{=}768$ in our model) using a single fully connected layer, and adding a learned positional embedding.
While both $\Tsep$ and $\Tjoint$ are stacks of multi-head self-attention (MHSA) layers,
$\Tcur$ uses masked ``conditional'' transformer layers, similar to Vaswani~\etal~\cite{vaswani2017attention}:
These alternate between masked MHSA layers and MHSA layers that use $\zjoint$ as keys (K) and values (V), as shown in Fig.~\ref{fig:details}. We use 6 transformer layers for $\Tsep$, 4 for $\Tjoint$, and 5 masked transformer layers for $\Tcur$. We use 16 attention heads everywhere.
We learn a separate temporal positional embedding to add to the input of $\Tjoint$.

\paragraph{Image encoder $E$, decoder $D$} 
The image encoder and decoder $E, D$ are not the focus of this paper, so we use architectures based on standard image compression approaches~\cite{minnen2018joint,minnen2020channel}. For the encoder, we use 4 strided convolutional layers, downscaling by a factor $16{\times}$ in total.
For the decoder, we use transposed convolutions and additionally add residual blocks at the low resolutions. 
We use $d_{ED}{=}192$ filters for all layers. See App.~\ref{sec:sweepD} for details and an exploration of architecture variants.

\begin{figure}
	{\footnotesize
	\hspace{1.9em}\textit{a) Shift} 
	\hspace{12em}\textit{b) SharpenOrBlur} 
	\hspace{7.0em}\textit{c) Fade}} 
	\\
    \includegraphics[width=\textwidth]{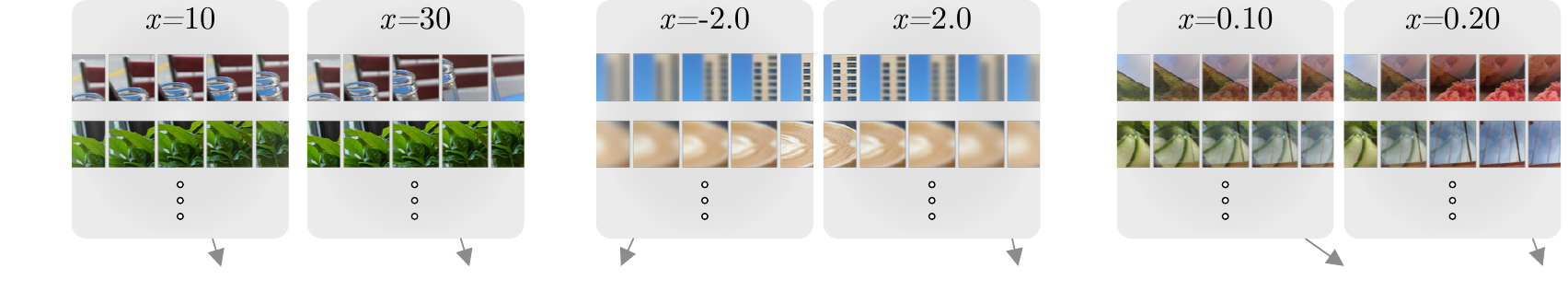} \\[-1.65ex]
	\includegraphics[width=\textwidth]{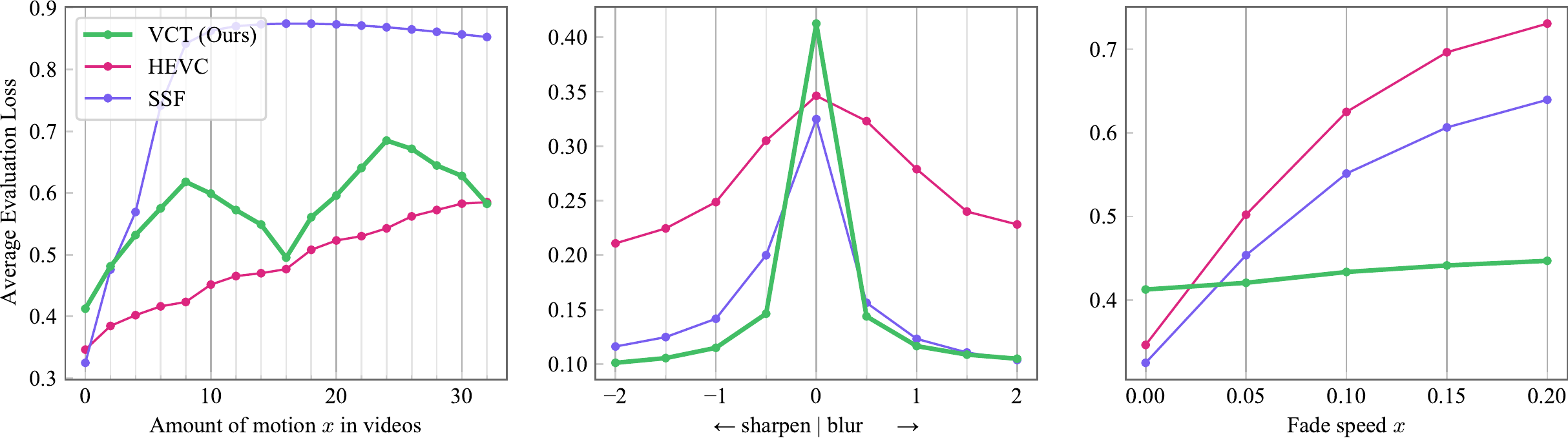}  \\[-3ex]
	\caption{
	\label{fig:synthetic} 
	To understand what types of temporal patterns our transformer has learned to exploit, we synthesize videos representing commonly seen patterns. 
	We compare to HEVC, which has built-in support for motion, and SSF, which has built-in support for motion and blurrying. VCT learns to handle all patterns purely from data. We refer to the text for a discussion.
	}
\end{figure}

\subsection{Loss and Training Process} 
\label{sec:loss_and_training}

The modeling choices introduced in the previous section allow for an efficient training procedure where we decompose the training into three stages, which enables rapid experimentation (Tab.~\ref{tab:training_setup}).
In \textbf{Stage I} we train the per-frame encoder $E$ and decoder $D$ by minimizing the rate-distortion trade-off~\cite[Sec 3.1.1]{Yang2022a}. Let $\mathcal U$ denote a uniform distribution in $[-0.5,0.5]$. We minimize
\begin{equation}
\mathcal{L}_\text{I} = \mathbb{E}_{%
  x\sim p_{X},
  u\sim\mathcal U
  }[
  \underbrace{-\text{log}\,p(\tilde y+u)}_{\text{bit-rate } r} +
  \lambda
  \underbrace{\text{MSE}(x, \hat x)}_{\text{distortion } d}], 
\qquad \tilde y{=}E(x),\, \hat x{=}D(\text{round}_\text{STE}(\tilde y)),
\label{eq:loss}
\end{equation}
using $\tilde y$ to refer to the unquantized representation, and $x \sim p_{X}$ are frames drawn from the training set.
Intuitively, we want to minimize the reconstruction error under the constraint that we can effectively quantize the encoder output, with $\lambda$ controlling the tradeoff.
For \textit{Stage I}, we thus employ the mean-scale hyperprior~\cite{minnen2018joint} approach to estimate $p$, the de facto standard in neural image compression, which we discard for later stages.\footnote{In short, the hyperprior estimates the PMF of $y$ using a VAE~\cite{kingma2013auto}, by predicting $p(y|z)$, where $z$ is side information transmitted first. We refer to the paper for details~\cite{minnen2018joint}.} 
To enable end-to-end training, we also follow~\cite{minnen2018joint}, 
adding i.i.d.\ uniform noise $u$ to $\tilde y$ when calculating $r$,
and using straight-through estimation (STE)~\cite{theis2017lossy,minnen2020channel} for gradients when rounding $\tilde y$ to feed it to $D$.

For \textbf{Stage II}, we train the transformer to obtain $p$, and only minimize rate:
\begin{equation}
\mathcal{L}_\text{II} = \mathbb{E}_{%
  (x_1,x_2,x_3)\sim p_{X_{1:3}},
  u\sim\mathcal U}[
  -\text{log}\,p(\tilde y_3+u|y_1,y_2)]
\qquad \tilde y_i{=}E(x),\, y_i{=}\text{round}(\tilde y_i),
\label{eq:loss2}
\end{equation}
where $(x_1,x_2,x_3)\sim p_{X_{1:3}}$ are triplets of adjacent video frames.
We assume each of the $\numc$ unquantized elements in each token follow a Gaussian distribution, $p\sim\mathcal N$,
and let the transformer predict $\numc$ means and $\numc$ scales per token.
Finally, we finetune everything jointly in \textbf{Stage III}, adding the distortion loss $d$ from Eq.~\ref{eq:loss} to Eq.~\ref{eq:loss2}.

We note that it is also possibe to train the model \textbf{from scratch} and obtain even better performance, see App.~\ref{sec:simplified}.

To obtain a discrete PMF $P$  for the quantized symbols (for entropy coding), we again follow standard practice~\cite{balle2018variational}, convolving $p$ with a unit-width box and evaluating it at discrete points, $P(y)=\int_{u \in \mathcal{U}}p(y+u) du,y\in\mathbb Z$~\cite[see, \eg,][Sec.~3.3.3, for details]{Yang2022a}.
To train, we use random spatio-temporal crops of $(B,N_F,256,256,3)$ pixels, where $B$ is the batch size, and $N_F$ the number of frames (values are given in Tab.~\ref{tab:training_setup}).
We use the linearly decaying learning rate  (LR) schedule with warmup, where we warmup for 10k steps and then linearly decay from the LR shown in the table to $1\textsc{e}^{-5}$.
\textit{Stage I} is trained using $\lambda{=}0.01$. To navigate the rate-distortion trade-off and obtain results for multiple rates, we fine-tune 9 models in \textit{Stage III}, using $\lambda{=}0.01 \cdot 2^i , i{\in}\{-3, \dots,5\}$.
We train all models on 4 Google Cloud TPUv4 chips.

\subsection{Latent Residual Predictor (LRP)} \label{sec:lrp}

To further leverage the powerful representation that the transformer learns, 
we adapt the ``latent residual predictor'' (LRP) from recent work in image compression~\cite{minnen2020channel}: The final features $\zcur$ from $\Tcur$  have the same spatial dimensions as $y_i$, and contain everything the transformer knows about the current and previous representations. 
Since we have to compute them to compute $P$, they constitute  ``free'' extra features that are helpful to reconstruct $\hat x_i$. 
We thus use $\zcur$ by feeding $y_i' = y_i + f_\text{LRP}(\zcur)$ to $D$ (we enable this in \textit{Stage III}), 
where $f_\text{LRP}$ consists of a $1{\times}1$ convolution mapping from $\dmodel$ to $d_{ED}$ followed by a residual block.
We note that this implies that $\hat x_i = D(y_i')$ indirectly depends on $y_{i-2}, y_{i-1}, y_i$. Since this is a bounded window into the past and $y_i'$ does not depend on $\hat{x}_{j < i}$, we remain free from temporal error propagation. 

\begin{figure}[t]
\centering
{\footnotesize
\begin{tabular}{@{\hskip 0mm}
                l@{\hskip 0.5mm}
                l@{\hskip 0.5mm}
                l@{\hskip 0.5mm}
                l@{\hskip 0.5mm}
                l%
                }
    &&\multicolumn{3}{@{\hskip 0mm}c@{\hskip 0mm}}{%
        \includegraphics[width=0.6\textwidth]{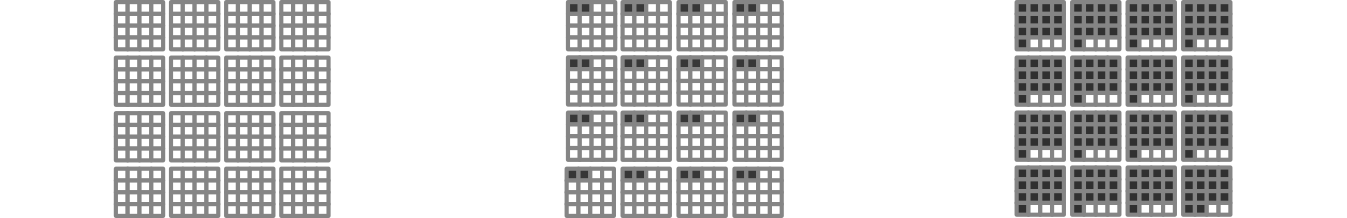}}\\
     \includegraphics[width=0.198\linewidth]{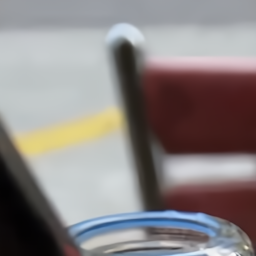} &
     \includegraphics[width=0.198\linewidth]{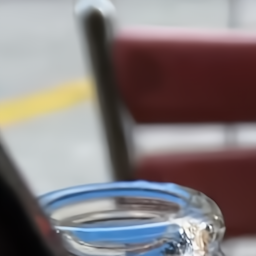} &
     \includegraphics[width=0.198\linewidth]{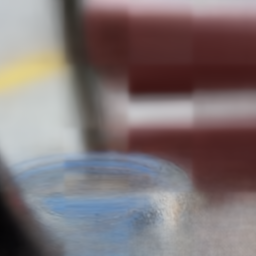} &
     \includegraphics[width=0.198\linewidth]{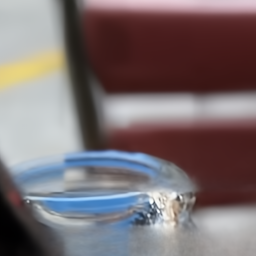} &
     \includegraphics[width=0.198\linewidth]{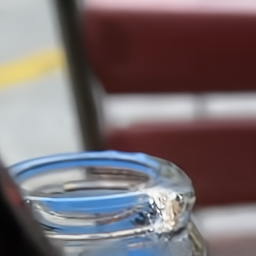} \\
     $\hat x_{i-2}$ &
     $\hat x_{i-1}$ &
      Decode: 
      \hfill  0$\tfrac{\text{tokens}}{\text{block}}$   &
      \hfill  2$\tfrac{\text{tokens}}{\text{block}}$  &
      \hfill 13$\tfrac{\text{tokens}}{\text{block}}$ \\[1mm]
      && \hfill 0kB\hspace{1pt}
      &  \hfill 0.13kB\hspace{1pt}
      &  \hfill 0.78kB\hspace{1pt}
      \\[-1ex]
\end{tabular}
}
\caption{\label{fig:sampling}
Visualizing the sample mean from the block-autoregressive distribution predicted by the transformer, as we decode more and more tokens (see Sec.~\ref{sec:results}).
We show the kilobytes (kB) required to transmit the decoded (gray) tokens.
On the left, we see the two previous reconstructions $\hat x_{i-2}, \hat x_{i-1}$.
In the middle, we see what the transformer expects at the current frame, \emph{before decoding any information} (0kB).
The next two images shows that  
as we decode more tokens, the model gets more certain, and the image obtained from the sample mean sharpens. Note that we never sample from the model for actual video coding.
}
\end{figure}

\section{Experiments}\label{sec:experiments}

\subsection{Data sets}
We train on one million Internet video clips, where each clip has nine frames. We obtained high-resolution videos which we downscale with a random factor (removing previous compression artifacts), from which we get a central 256 crop. 
Training batches are made up of randomly selected triplets of adjacent frames.
We evaluate on two common benchmark data sets: (1) \textit{MCL-JCV}~\cite[MIT Licence]{wang2016mcl} made up of thirty 1080p videos captured at either 25 or 30FPS and averaging $137$ frames per video, 
and (2) \textit{UVG}~\cite[CC-BY-NC Licence]{mercat2020uvg} containing twelve 1080p 120FPS videos with either 300 or 600 frames each.

\paragraph{Synthetic videos}
We explore three parameterized synthetic data sets that we build by generating videos from still images from the CLIC2020 test set~\cite[Unsplash licence]{clic2020},
(see Fig.~\ref{fig:synthetic}). 
Each data set has a parameter $x$ that we vary, and we create 100 videos for each value of $x$. Each video is 12 frames of $512{\times}512$px. We explore:
\textbf{Shift}, where we pan from the center of the image towards the lower right, shifting by $x$ pixels in each step.
\textbf{SharpenOrBlur}, where if $x{\geq}0$, we apply Gaussian blurring with sigma $x\cdot t$ at time step $t$. If $x{<}0$, we create videos that get sharper over time by playing a video blurred with $|x|$ in reverse.
\textbf{Fade}, where we linearly transition between two unrelated images using alpha blending (as in a scene cut). We release the code to synthesize these videos.

\begin{table}[t]
    \centering
    \begin{tabular}{lccll}
    \toprule
      & Context & LRP & bpp $\downarrow$ & PSNR $\uparrow$ \\
      \midrule
      No previous frames (image codec) 
                            & 0 & & 0.218 & 36.1 \\
      \midrule
      1 previous frame      & 1 & & 0.0907 (-58\%) & 36.1 \\
      2 previous frames     & 2 & & 0.0775 (-64\%) & 36.1 \\
      2 previous frames and LRP (VCT (Ours))
                                & 2 & \checkmark & 0.0775 (-64\%) & 36.8 (+0.7dB) \\
      \bottomrule \\[-1ex]
    \end{tabular}
    \caption{Ablating how many previous frames we feed to the transformer (``Context''), and whether we use latent-residual prediction (LRP).
    \label{tab:context}
    }
\end{table}

\subsection{Models}
We refer to our video compression transformer as \textbf{\name}. 
We run the widely used, non-neural, standard codec \textbf{HEVC}~\cite{sullivan2012overview} (\textit{a.k.a.} H.265) using the ffmpeg x265 codec in the \textit{medium} and \textit{veryslow} settings,
as well as \textbf{H.264} using x264 in the \textit{medium} setting.
For a fair comparison to our method, we follow previous work~\cite{agustsson2020scale,mentzer2021neural,rippel2021elf} in disabling B-Frames,
but do not constrain the codecs in any other way. 
We run the public \textbf{DVC}~\cite{lu2019dvc} code, and additionally obtain numbers from the following papers:
\textbf{SSF}~\cite{agustsson2020scale}, which introduced scale-space-flow, an architectural component to support warping and blurring, commonly used in follow-up work,
\textbf{ELF-VC}~\cite{rippel2021elf}, to the best of our knowledge the state-of-the-art neural method in terms of PSNR on MCL-JCV, which extends the motion compensation of SSF with more motion priors,
\textbf{FVC}~\cite{hu2021fvc} and
\textbf{DCVC}~\cite{li2021deep},
both strong models based on warping plus residual coding in a representation space,
\textbf{RLVC}~\cite{yang2021rlvc}, using ConvLSTMs as a sequence model,
and \textbf{Liu~\etal}~\cite{liu2020conditional}, who study losslessly transmitting representations using CNNs for temporal entropy modelling.
To explore how architectural biases behave on synthetic data, 
we reproduce SSF, using exactly the same training data as for VCT.

\subsection{Metrics}

We evaluate the common PSNR and MS-SSIM~\cite{wang2003multiscale} in RGB. We train all models using MSE as a distortion and use $200\cdot(1 - \text{MS-SSIM}(x,\hat x))$ as the training objective in \textit{Stage III} (Tab.~\ref{tab:training_setup}) to obtain MS-SSIM models.

\section{Results} \label{sec:results}

\subsection{Comparison to State of the Art}
In Fig.~\ref{fig:rd_results}, we depict rate distortion graphs for our method and the neural video compression methods introduced in Sec.~\ref{sec:experiments}, on MCL-JCV and UVG.
Despite the simplicity of our approach, and the fact that we use no motion or warping components, we outperform all methods in both PSNR and MS-SSIM.

\subsection{Synthetic data}
In Fig.~\ref{fig:synthetic}, we show how the transfomer learns to exploit various types of temporal patterns by applying it to the synthetic data sets introduced in Sec.~\ref{sec:experiments}, and reporting the evaluation R-D loss.\footnote{%
$\mathcal{L}{=}r + \lambda d$. To calculate $\mathcal{L}$ for HEVC, we find the quality factor $q$ matching our $\lambda$ via $q{=}\text{arg\;min}_q r(q)+\lambda d(q)$, which yields $q{=}25$ for $\lambda{=}0.01$.
}
We compare to HEVC and SSF, which both have explicit support for shifting motion, while SSF also has explicit support for blurring. We expect them to perform well on temporal patterns for which they have corresponding architectural priors. 
In contrast, VCT has no such priors.
For each data set, we explore different values for the parameter $x$ (see Sec.~\ref{sec:experiments}), a point in the plot represents the average evaluation loss over the 100 videos created with $x$. 

We observe: a) 
On videos with shifting based motion, VCT obtains ${\approx}45\%$ lower R-D loss compared to SSF, which saturates at about $x=10$, presumably due to the shallow CNN used for flow estimation.
Since HEVC supports motion compensating with arbitrary shifts of previous frames, it excels on these kinds of videos.
For shifts that are a multiple of 16, the representations shifts by exactly 1 symbol in each step, and VCT matches HEVC.
The reason for this is that our encoder is a CNN, so it is only shift-equivariant for shifts which are multiples of the stride (16). 
Any shift in [1, 15] pixels causes the representation to change in a complex way (cf.~\cite{zhang2019making}).
b) For blurring/sharpening, we outperform both HEVC and SSF, despite the latter having explicit support for blurring. 
Note that the curve for SSF is asymmetric: since it has built-in support for blurring, it gets a ${\approx}20\%$ lower RD loss on blurring compared to sharpening.
c) VCT learns to handle fading,
exhibiting a near-constant RD loss as we increase $x$, in contrast to the baselines, neither of which has a explicit support for fading. SSF is ${\approx}20\%$ better than HEVC, possibly due to its blurrying capabilities.
For completely static videos $x{=}0$, we observe that VCT is at a slight disadvantage compared to the previous approaches. 
Overall, we believe that synthetic data can give better insight into the strengths and weaknesses of methods, and hope that future work can compare on these data sets.

\subsection{Visualizing certainty during decoding}
After having seen $k$ tokens in each block, the transformer predicts a PMF $P(t_{k+1}|t_{\leq k},\zjoint)$.
This induces a joint distribution $P(t_{>k}|\dots)$ over all unseen (not yet decoded) tokens. 
Intuitively, if the transformer is certain about the future, this distribution will be concentrated on the actual future tokens we will decode.
In Fig.~\ref{fig:sampling}, we visualize the \textit{sample mean} of this distribution by feeding it through $D$, \ie, we sample $N$ realisations of the unseen tokens in each block, conditioned on the $k$ already decoded ones, for $k \in \{0, 2, 13\}$.
In the middle image in Fig.~\ref{fig:sampling}, we show what the transformer expects at the current frame, 
\emph{before decoding any information} ($k=0$, \ie, 0 bits).
We observe that the model is able---to some degree---to learn second order motion implicitly.
The next two images shows that as we decode more tokens, the model gets more certain, and the image sharpens.

\subsection{Ablations}
In Tab.~\ref{tab:context}, we explore 
the importance of temporal context from previous frames and latent residual prediction (LRP) on MCL-JCV.
We start from a baseline that does not use any 
previous frames, \ie, an image model, used to independently code each frame.
Conditioning on one previous frame reduces bitrate by $-58\%$.
Using two previous frames yields an additional improvement of $-6\%$.
In the final configuration (our model, VCT), which adds LRP, we observe an increase in PSNR of 0.7dB at the same bitrate. We did not observe further gains from more context.

\begin{table}[t]
    \centering
    \begin{tabular}{%
        llcc@{\hskip 2em}c@{\hskip  2em}cr
    }
    \toprule
         &
         & $\Tsep$ and $\Tjoint$
         & $\Tcur$
         & EC
         & $D$
         & FPS estimate \\
    \midrule
         Ours
         &   1080p & \phantom{.}168 ms &  \phantom{.}326 ms & 30.5\phantom{0} ms & 168 ms & $\approx$1.4 FPS\\
         &    720p  & 37.6 ms &  44.8 ms & 17.0\phantom{0} ms & 49.5 ms & $\approx$6.7 FPS\\
         &    480p  & 18.1 ms &  23.1 ms & \phantom{1}9.02 ms & 23.3 ms & $\approx$13.6 FPS  \\
         &    360p  & 
       \phantom{1}7.3 ms &  14.9 ms & \phantom{1}4.24 ms & 10.1 ms & $\approx$27.3 FPS \\
    \bottomrule \\[-1ex]
    \end{tabular}
    \caption{
    \label{tab:decoding_speed}
    Runtimes of our components. For ours, we use a Google Cloud TPU v4 to run transformers and $D$. Entropy Coding (EC) is run on CPU.
    }
\end{table}

\subsection{Runtime}
To obtain runtimes of the transformers ($\Tsep, \Tjoint, \Tcur$) and the decoder ($D$), we employ a Google Cloud TPU v4 (single core) using Flax~\cite{flax2020github}, which has an efficient implementation for autoregressive transformers.
\begin{table}[b]
    \centering
    \begin{tabular}{%
        llr
    }
    \toprule
         & Resolution & FPS estimate \\
    \midrule
         Ours
         &   1080p  &  $\approx$1.4 FPS\\
         &    720p  &  $\approx$6.7 FPS\\
         &    480p  &  $\approx$13.6 FPS  \\
         &    360p  &  $\approx$27.3 FPS \\
   \midrule
        DCVC~\cite{li2021deep} 
             & 1080p & $\approx$1.1 FPS \\
   \midrule
        FVC~\cite{hu2021fvc}
             & 1080p & $\approx$1.8 FPS \\
   \midrule
        ELF-VC~\cite{rippel2021elf}
             & 1080p & $\approx$18 FPS \\
             & 720p & $\approx$35 FPS \\
    \bottomrule \\[-1ex]
    \end{tabular}
    \caption{
    \label{tab:decoding_speed2}Comparing decoding speed to other methods. We directly copy reported results from the respective papers, so platforms are not comparable.}
\end{table}
We use Tensorflow Compression to measure time spent entropy coding (EC), on an Intel Skylake CPU core. 
In Tab.~\ref{tab:decoding_speed}, we report numbers for $1280{\times}720$px, $852{\times}480$px, and $480{\times}360$px.
Since this benchmark is not fully end-to-end, we only report an FPS estimate by calculating $1000/(\text{sum of individual runtimes in ms})$.
Note that running $\Tcur$ at 720p once only takes ${\approx}2.8$ms, but we run it $w_\cur^2{=}16$ times to decode a frame.
To run $\Tjoint$, we only have to run $\Tsep$ once per representation, since we can re-use the output of running $\Tsep$ on the previous representation.

Many neural compression methods do not detail inference time and do not have code available, but we copy the results from 
DCVC~\cite{li2021deep},
FVC~\cite{hu2021fvc},
and ELF-VC~\cite{rippel2021elf}, 
in Table~\ref{tab:decoding_speed2}.

\section{Conclusion and Future Work}
\label{sec:conclusions}

We presented an elegantly simple transformer-based approach to neural video compression, outperforming previous methods without relying on architectural priors such as explicit motion prediction or warping.
Notably, our results are achieved by conditioning the transformer only on a 2-frame window into the past. For some types of videos, it would be interesting to scale this up, or to introduce a notion of more long-term memory, possibly leveraging arbitrary reference frames. 

As mentioned towards the end of Sec.~\ref{sec:overview}, various different ways to factorize the distributions could be explored, including vector quantization, channel-autoregression, or changing the independence assumptions around how we split representations into blocks.

\paragraph{Societal Impact}
We hope our method can serve as the foundation for a new generation of video codecs.
This could have a net-positive impact on society by reducing the bandwidth needed for video conferencing and video streaming and to better utilize storage space, therefore increasing the capacity of knowledge preservation.

\paragraph{Acknowledgements} We thank Basil Mustafa, Ashok Popat, Huiwen Chang, Phil Chou, Johannes Ball\'{e}, and Nick Johnston for the insightful discussions and feedback.

\newpage

{\footnotesize
\printbibliography

@inproceedings{wang2021max,
  title={Max-deeplab: End-to-end panoptic segmentation with mask transformers},
  author={Wang, Huiyu and Zhu, Yukun and Adam, Hartwig and Yuille, Alan and Chen, Liang-Chieh},
  booktitle={Proceedings of the IEEE/CVF Conference on Computer Vision and Pattern Recognition},
  pages={5463--5474},
  year={2021}
}

@inproceedings{zheng2021rethinking,
  title={Rethinking semantic segmentation from a sequence-to-sequence perspective with transformers},
  author={Zheng, Sixiao and Lu, Jiachen and Zhao, Hengshuang and Zhu, Xiatian and Luo, Zekun and Wang, Yabiao and Fu, Yanwei and Feng, Jianfeng and Xiang, Tao and Torr, Philip HS and others},
  booktitle={Proceedings of the IEEE/CVF conference on computer vision and pattern recognition},
  pages={6881--6890},
  year={2021}
}

@inproceedings{bertasius2021space,
  title={Is Space-Time Attention All You Need for Video Understanding?},
  author={Bertasius, Gedas and Wang, Heng and Torresani, Lorenzo},
  booktitle={International Conference on Machine Learning},
  pages={813--824},
  year={2021},
  organization={PMLR}
}

@inproceedings{neimark2021video,
  title={Video transformer network},
  author={Neimark, Daniel and Bar, Omri and Zohar, Maya and Asselmann, Dotan},
  booktitle={Proceedings of the IEEE/CVF International Conference on Computer Vision},
  pages={3163--3172},
  year={2021}
}

@inproceedings{sun2019videobert,
  title={Videobert: A joint model for video and language representation learning},
  author={Sun, Chen and Myers, Austin and Vondrick, Carl and Murphy, Kevin and Schmid, Cordelia},
  booktitle={Proceedings of the IEEE/CVF International Conference on Computer Vision},
  pages={7464--7473},
  year={2019}
}

@inproceedings{fan2021multiscale,
  title={Multiscale vision transformers},
  author={Fan, Haoqi and Xiong, Bo and Mangalam, Karttikeya and Li, Yanghao and Yan, Zhicheng and Malik, Jitendra and Feichtenhofer, Christoph},
  booktitle={Proceedings of the IEEE/CVF International Conference on Computer Vision},
  pages={6824--6835},
  year={2021}
}

@article{devlin2018bert,
  title={Bert: Pre-training of deep bidirectional transformers for language understanding},
  author={Devlin, Jacob and Chang, Ming-Wei and Lee, Kenton and Toutanova, Kristina},
  journal={arXiv preprint arXiv:1810.04805},
  year={2018}
}

@article{edunov2018understanding,
  title={Understanding back-translation at scale},
  author={Edunov, Sergey and Ott, Myle and Auli, Michael and Grangier, David},
  journal={arXiv preprint arXiv:1808.09381},
  year={2018}
}

@article{brown2020language,
  title={Language models are few-shot learners},
  author={Brown, Tom and Mann, Benjamin and Ryder, Nick and Subbiah, Melanie and Kaplan, Jared D and Dhariwal, Prafulla and Neelakantan, Arvind and Shyam, Pranav and Sastry, Girish and Askell, Amanda and others},
  journal={Advances in neural information processing systems},
  volume={33},
  pages={1877--1901},
  year={2020}
}

@article{vaswani2017attention,
  title={Attention is all you need},
  author={Vaswani, Ashish and Shazeer, Noam and Parmar, Niki and Uszkoreit, Jakob and Jones, Llion and Gomez, Aidan N and Kaiser, {\L}ukasz and Polosukhin, Illia},
  journal={Advances in neural information processing systems},
  volume={30},
  year={2017}
}

@inproceedings{arnab2021vivit,
  title={Vivit: A video vision transformer},
  author={Arnab, Anurag and Dehghani, Mostafa and Heigold, Georg and Sun, Chen and Lu{\v{c}}i{\'c}, Mario and Schmid, Cordelia},
  booktitle={Proceedings of the IEEE/CVF International Conference on Computer Vision},
  pages={6836--6846},
  year={2021}
}

@article{dosovitskiy2020image,
  title={An image is worth 16x16 words: Transformers for image recognition at scale},
  author={Dosovitskiy, Alexey and Beyer, Lucas and Kolesnikov, Alexander and Weissenborn, Dirk and Zhai, Xiaohua and Unterthiner, Thomas and Dehghani, Mostafa and Minderer, Matthias and Heigold, Georg and Gelly, Sylvain and others},
  journal={arXiv preprint arXiv:2010.11929},
  year={2020}
}

@inproceedings{liu2021swin,
  title={Swin transformer: Hierarchical vision transformer using shifted windows},
  author={Liu, Ze and Lin, Yutong and Cao, Yue and Hu, Han and Wei, Yixuan and Zhang, Zheng and Lin, Stephen and Guo, Baining},
  booktitle={Proceedings of the IEEE/CVF International Conference on Computer Vision},
  pages={10012--10022},
  year={2021}
}

@article{qian2022entroformer,
  title={Entroformer: A Transformer-based Entropy Model for Learned Image Compression},
  author={Qian, Yichen and Lin, Ming and Sun, Xiuyu and Tan, Zhiyu and Jin, Rong},
  journal={arXiv preprint arXiv:2202.05492},
  year={2022}
}

@inproceedings{zhu2021transformer,
  title={Transformer-based Transform Coding},
  author={Zhu, Yinhao and Yang, Yang and Cohen, Taco},
  booktitle={International Conference on Learning Representations},
  year={2021}
}

@article{chowdhery2022palm,
  title={Palm: Scaling language modeling with pathways},
  author={Chowdhery, Aakanksha and Narang, Sharan and Devlin, Jacob and Bosma, Maarten and Mishra, Gaurav and Roberts, Adam and Barham, Paul and Chung, Hyung Won and Sutton, Charles and Gehrmann, Sebastian and others},
  journal={arXiv preprint arXiv:2204.02311},
  year={2022}
}

@inproceedings{wang2003multiscale,
  title={Multiscale structural similarity for image quality assessment},
  author={Wang, Zhou and Simoncelli, Eero P and Bovik, Alan C},
  booktitle={The Thrity-Seventh Asilomar Conference on Signals, Systems \& Computers, 2003},
  volume={2},
  pages={1398--1402},
  year={2003},
  organization={Ieee}
}

@article{yang2021perceptual,
  title={Perceptual Learned Video Compression with Recurrent Conditional GAN},
  author={Yang, Ren and Van Gool, Luc and Timofte, Radu},
  journal={arXiv preprint arXiv:2109.03082},
  year={2021}
}

@article{mentzer2021neural,
  title={Neural Video Compression using GANs for Detail Synthesis and Propagation},
  author={Mentzer, Fabian and Agustsson, Eirikur and Ball{\'e}, Johannes and Minnen, David and Johnston, Nick and Toderici, George},
  journal={arXiv preprint arXiv:2107.12038},
  year={2021}
}

@inproceedings{golinski2020feedback,
  title={Feedback recurrent autoencoder for video compression},
  author={Golinski, Adam and Pourreza, Reza and Yang, Yang and Sautiere, Guillaume and Cohen, Taco S},
  booktitle={Proceedings of the Asian Conference on Computer Vision},
  year={2020}
}

@article{yang2021rlvc,
  title={Learning for Video Compression with Recurrent Auto-Encoder and Recurrent Probability Model},
  author={Yang, Ren and Mentzer, Fabian and Van Gool, Luc and Timofte, Radu},
  journal={IEEE Journal of Selected Topics in Signal Processing},
  volume={15},
  number={2},
  pages={388-401},
  year={2021}
}

@article{li2021deep,
  title={Deep contextual video compression},
  author={Li, Jiahao and Li, Bin and Lu, Yan},
  journal={Advances in Neural Information Processing Systems},
  volume={34},
  year={2021}
}

@inproceedings{liu2020conditional,
  title={Conditional entropy coding for efficient video compression},
  author={Liu, Jerry and Wang, Shenlong and Ma, Wei-Chiu and Shah, Meet and Hu, Rui and Dhawan, Pranaab and Urtasun, Raquel},
  booktitle={European Conference on Computer Vision},
  pages={453--468},
  year={2020},
  organization={Springer}
}

@article{yang2020hierarchical,
  title={Hierarchical autoregressive modeling for neural video compression},
  author={Yang, Ruihan and Yang, Yibo and Marino, Joseph and Mandt, Stephan},
  journal={arXiv preprint arXiv:2010.10258},
  year={2020}
}

@inproceedings{liu2021deep,
  title={Deep learning in latent space for video prediction and compression},
  author={Liu, Bowen and Chen, Yu and Liu, Shiyu and Kim, Hun-Seok},
  booktitle={Proceedings of the IEEE/CVF Conference on Computer Vision and Pattern Recognition},
  pages={701--710},
  year={2021}
}

@inproceedings{habibian2019video,
  title={Video compression with rate-distortion autoencoders},
  author={Habibian, Amirhossein and Rozendaal, Ties van and Tomczak, Jakub M and Cohen, Taco S},
  booktitle={Proceedings of the IEEE/CVF International Conference on Computer Vision},
  pages={7033--7042},
  year={2019}
}

@inproceedings{lu2019dvc,
  title={Dvc: An end-to-end deep video compression framework},
  author={Lu, Guo and Ouyang, Wanli and Xu, Dong and Zhang, Xiaoyun and Cai, Chunlei and Gao, Zhiyong},
  booktitle={Proceedings of the IEEE/CVF Conference on Computer Vision and Pattern Recognition},
  pages={11006--11015},
  year={2019}
}

@inproceedings{wu2018video,
  title={Video compression through image interpolation},
  author={Wu, Chao-Yuan and Singhal, Nayan and Krahenbuhl, Philipp},
  booktitle={Proceedings of the European Conference on Computer Vision (ECCV)},
  pages={416--431},
  year={2018}
}

@inproceedings{djelouah2019neural,
  title={Neural inter-frame compression for video coding},
  author={Djelouah, Abdelaziz and Campos, Joaquim and Schaub-Meyer, Simone and Schroers, Christopher},
  booktitle={Proceedings of the IEEE/CVF International Conference on Computer Vision},
  pages={6421--6429},
  year={2019}
  }

@unpublished{Yang2022a,
  author = "Y. Yang and S. Mandt and L. Theis",
  title = "An Introduction to Neural Data Compression",
  year = 2022,
  url = "https://arxiv.org/abs/2202.06533",
  note = "preprint"
}

@inproceedings{agustsson2020scale,
  title={Scale-space flow for end-to-end optimized video compression},
  author={Agustsson, Eirikur and Minnen, David and Johnston, Nick and Balle, Johannes and Hwang, Sung Jin and Toderici, George},
  booktitle={Proceedings of the IEEE/CVF Conference on Computer Vision and Pattern Recognition},
  pages={8503--8512},
  year={2020}
}

@inproceedings{wang2016mcl,
  title={{MCL-JCV}: a {JND}-based {H.264/AVC} video quality assessment dataset},
  author={Wang, Haiqiang and Gan, Weihao and Hu, Sudeng and Lin, Joe Yuchieh and Jin, Lina and Song, Longguang and Wang, Ping and Katsavounidis, Ioannis and Aaron, Anne and Kuo, C-C Jay},
  booktitle={2016 IEEE International Conference on Image Processing (ICIP)},
  pages={1509--1513},
  year={2016},
  organization={IEEE}
}

@inproceedings{mercat2020uvg,
  title={{UVG dataset: 50/120fps 4K sequences for video codec analysis and development}},
  author={Mercat, Alexandre and Viitanen, Marko and Vanne, Jarno},
  booktitle={Proceedings of the 11th ACM Multimedia Systems Conference},
  pages={297--302},
  year={2020}
}

@inproceedings{hu2021fvc,
  title={FVC: A new framework towards deep video compression in feature space},
  author={Hu, Zhihao and Lu, Guo and Xu, Dong},
  booktitle={Proceedings of the IEEE/CVF Conference on Computer Vision and Pattern Recognition},
  pages={1502--1511},
  year={2021}
}

@inproceedings{zhang2019making,
  title={Making convolutional networks shift-invariant again},
  author={Zhang, Richard},
  booktitle={International conference on machine learning},
  pages={7324--7334},
  year={2019},
  organization={PMLR}
}

@article{balle2020nonlinear,
  title={Nonlinear transform coding},
  author={Ball{\'e}, Johannes and Chou, Philip A and Minnen, David and Singh, Saurabh and Johnston, Nick and Agustsson, Eirikur and Hwang, Sung Jin and Toderici, George},
  journal={IEEE Journal of Selected Topics in Signal Processing},
  volume={15},
  number={2},
  pages={339--353},
  year={2020},
  publisher={IEEE}
}

@article{minnen2020channel,
  title={Channel-wise autoregressive entropy models for learned image compression},
  author={Minnen, David and Singh, Saurabh},
  journal={arXiv preprint arXiv:2007.08739},
  year={2020}
}

@article{sullivan2012overview,
  title={Overview of the high efficiency video coding (HEVC) standard},
  author={Sullivan, Gary J and Ohm, Jens-Rainer and Han, Woo-Jin and Wiegand, Thomas},
  journal={IEEE Transactions on circuits and systems for video technology},
  volume={22},
  number={12},
  pages={1649--1668},
  year={2012},
  publisher={IEEE}
}

@inproceedings{balle2018variational,
    Author = {Ball{\'e}, Johannes and Minnen, David and Singh, Saurabh and Hwang, Sung Jin and Johnston, Nick},
    Booktitle = {International Conference on Learning Representations (ICLR)},
    Title = {Variational image compression with a scale hyperprior},
    Year = {2018}}

@inproceedings{minnen2018joint,
    title={Joint autoregressive and hierarchical priors for learned image compression},
    author={Minnen, David and Ball{\'e}, Johannes and Toderici, George D},
    booktitle={Advances in Neural Information Processing Systems},
    pages={10771--10780},
    year={2018}
}

@inproceedings{theis2017lossy,
    Author = {Theis, Lucas and Shi,Wenzhe and Cunningham, Andrew and Huszar, Ferenc},
    Booktitle = {International Conference on Learning Representations (ICLR)},
    Title = {Lossy image compression with compressive autoencoders},
    Year = {2017}}

@article{kingma2013auto,
    Author = {Kingma, Diederik P and Welling, Max},
    Journal = {arXiv preprint arXiv:1312.6114},
    Title = {Auto-encoding variational bayes},
    Year = {2013}}

@inproceedings{rippel2021elf,
  title={Elf-vc: Efficient learned flexible-rate video coding},
  author={Rippel, Oren and Anderson, Alexander G and Tatwawadi, Kedar and Nair, Sanjay and Lytle, Craig and Bourdev, Lubomir},
  booktitle={Proceedings of the IEEE/CVF International Conference on Computer Vision},
  pages={14479--14488},
  year={2021}
}

@manual{clic2020,
    title={CLIC 2020: Challenge on Learned Image Compression},
    author={George Toderici and Lucas Theis and Nick Johnston and Eirikur Agustsson and
    Fabian Mentzer and Johannes Ball{\'e} and Wenzhe Shi and Radu Timofte},
    note="\url{http://compression.cc}",
    year={2020},
}

@misc{flax2020github,
  author = {Jonathan Heek and Anselm Levskaya and Avital Oliver and Marvin Ritter and Bertrand Rondepierre and Andreas Steiner and Marc van {Z}ee},
  title = {{F}lax: A neural network library and ecosystem for {JAX}},
  url = {http://github.com/google/flax},
  version = {0.4.2},
  year = {2020},
}

@inproceedings{he2022elic,
  title={Elic: Efficient learned image compression with unevenly grouped space-channel contextual adaptive coding},
  author={He, Dailan and Yang, Ziming and Peng, Weikun and Ma, Rui and Qin, Hongwei and Wang, Yan},
  booktitle={Proceedings of the IEEE/CVF Conference on Computer Vision and Pattern Recognition},
  pages={5718--5727},
  year={2022}
}
}

 \newpage
 \section*{NeurIPS Checklist}

 \begin{enumerate}

 \item For all authors...
 \begin{enumerate}
   \item Do the main claims made in the abstract and introduction accurately reflect the paper's contributions and scope?
     \answerYes{}
   \item Did you describe the limitations of your work?
     \answerYes{see Sec~\ref{sec:conclusions}.}
   \item Did you discuss any potential negative societal impacts of your work?
     \answerYes{see Societal Impact in Sec~\ref{sec:conclusions}.}
   \item Have you read the ethics review guidelines and ensured that your paper conforms to them?
     \answerYes{}
 \end{enumerate}

 \item If you are including theoretical results...
 \begin{enumerate}
   \item Did you state the full set of assumptions of all theoretical results?
     \answerNA{}
         \item Did you include complete proofs of all theoretical results?
     \answerNA{}
 \end{enumerate}

 \item If you ran experiments...
 \begin{enumerate}
   \item Did you include the code, data, and instructions needed to reproduce the main experimental results (either in the supplemental material or as a URL)?
     \answerNo{We cannot release training data but will release code if the paper is published.}
   \item Did you specify all the training details (e.g., data splits, hyperparameters, how they were chosen)?
     \answerYes{}
         \item Did you report error bars (e.g., with respect to the random seed after running experiments multiple times)?
     \answerNo{However, we find in most experiments, multiple runs end at similar final losses.}
         \item Did you include the total amount of compute and the type of resources used (e.g., type of GPUs, internal cluster, or cloud provider)?
     \answerYes{We specify training platform and training times in \ref{sec:loss_and_training}, as well as how many models we train.}
 \end{enumerate}

 \item If you are using existing assets (e.g., code, data, models) or curating/releasing new assets...
 \begin{enumerate}
   \item If your work uses existing assets, did you cite the creators?
     \answerYes{See Sec~\ref{sec:experiments}.}
   \item Did you mention the license of the assets?
     \answerYes{See Sec~\ref{sec:experiments}.}
   \item Did you include any new assets either in the supplementary material or as a URL?
      \answerYes{We will release a Github URL to our code upon publication.}
   \item Did you discuss whether and how consent was obtained from people whose data you're using/curating?
     \answerNA{We don't release new data.}
   \item Did you discuss whether the data you are using/curating contains personally identifiable information or offensive content?
     \answerNA{We don't release new data.}
 \end{enumerate}

 \item If you used crowdsourcing or conducted research with human subjects...
 \begin{enumerate}
   \item Did you include the full text of instructions given to participants and screenshots, if applicable?
     \answerNA{No crowdsourcing or human subjects.}
   \item Did you describe any potential participant risks, with links to Institutional Review Board (IRB) approvals, if applicable?
     \answerNA{No crowdsourcing or human subjects.}
   \item Did you include the estimated hourly wage paid to participants and the total amount spent on participant compensation?
     \answerNA{No crowdsourcing or human subjects.}
 \end{enumerate}

 \end{enumerate}

\newpage

\appendix

\section*{VCT: A Video Compression Transformer -- Supplementary Material}

\section{Appendix}

\subsection[Main Text Image auto-encoder]{Main Text Image auto-encoder ($E,D$) details} \label{sec:sweepD}

For the results of the main text we use the following architectures for $E,D$.
Let \texttt{C} be a $5{\times}5$ convolution with $d_{ED}=192$ filters and stride 2, followed by a leaky relu activation (with $\alpha=0.2$). Our encoder $E$ is \texttt{CCCC}. Let \texttt{T} be a $5{\times}5$ transposed convolution with $d_{ED}$ filters and stride 2, also followed by a leaky relu, and let \texttt{R} be a residual block (\ie, \texttt{R} is \texttt{CC} with a skip connection around it).
$D$ is \texttt{RRRRTRRTRRTT}, \ie, as we increase in resolution we use fewer residual blocks. 

\begin{figure}[h]
    \centering
    \includegraphics[width=0.5\textwidth]{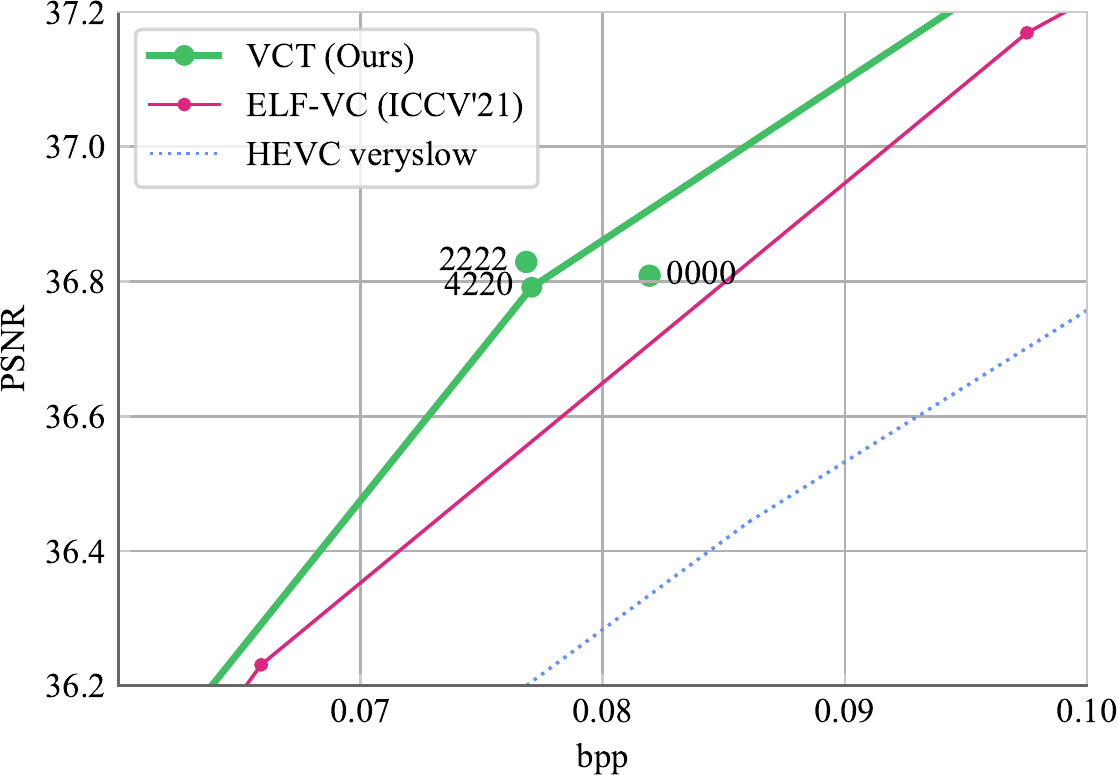}
    \caption{\label{fig:sweepD}Exploring variants for the decoder $D$, on MCL-JCV.}
\end{figure}

We use the shorthand $4220$ for this, counting the residual blocks between each transpose convolution \texttt{T}. 
In Fig.~\ref{fig:sweepD}, we explore $0000$ (no residual blocks) and $2222$. 
The latter has the same number of residual blocks as our defaults, but uses them in a later stage, making them more expensive (high resolution features). 

\subsection[Image auto-encoder]{Public Code Release: Simplified Training from Scratch} \label{sec:simplified}

\begin{figure}[h]
    \centering
    \includegraphics[width=0.5\textwidth]{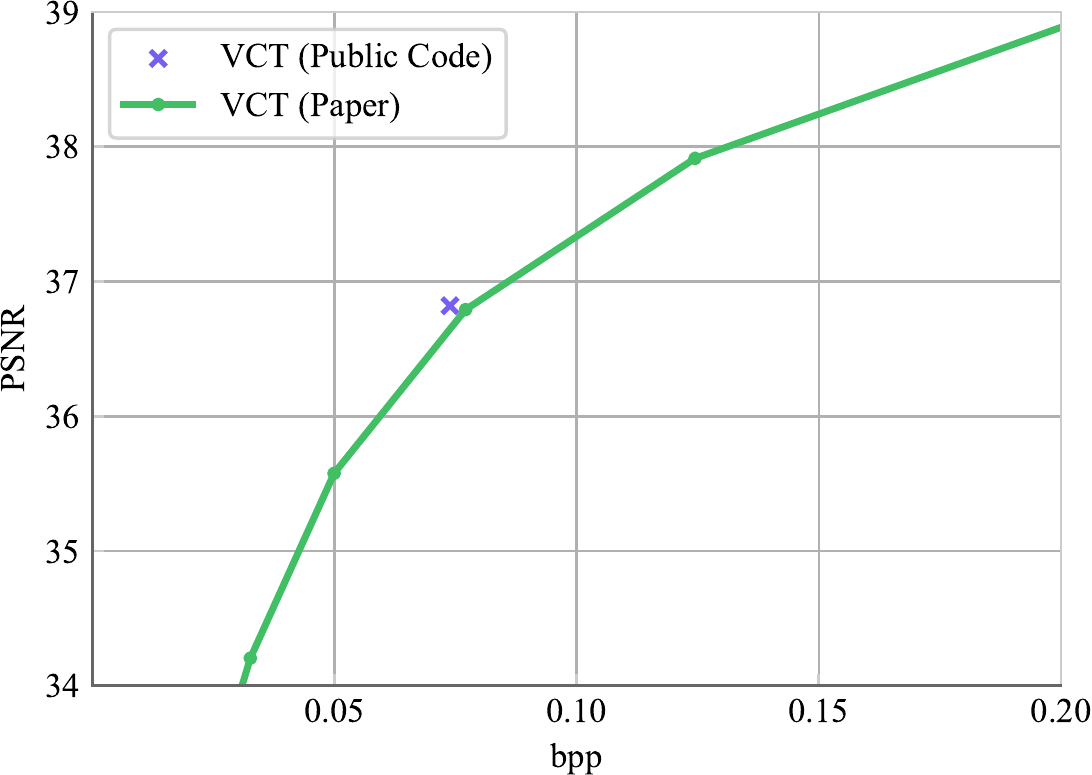}
    \caption{\label{fig:public}Performance when training from scratch with the public code, on MCL-JCV.}
\end{figure}

The official code release at \url{https://goo.gle/vct-paper} contains a simplified training setup. We only train Stage III (Table.~\ref{tab:training_setup}), directly \textbf{from scratch}, using a LR of $1\textsc{e}^{-4}$ for 750k steps. We find that the main $E,D$ (see Section~\ref{sec:sweepD}) are leading to unstable training when trained from scratch, so we instead use a light-weight architecture from ELIC~\cite{he2022elic}. The resulting model actually outperforms the architectures presented in the main text, see Fig.~\ref{fig:public}. We note that we do not need to train a Hyperprior in this setup.

\subsection{Training Set Size} \label{sec:app:data}
\begin{figure}[h]
    \centering
    \includegraphics[width=0.5\linewidth]{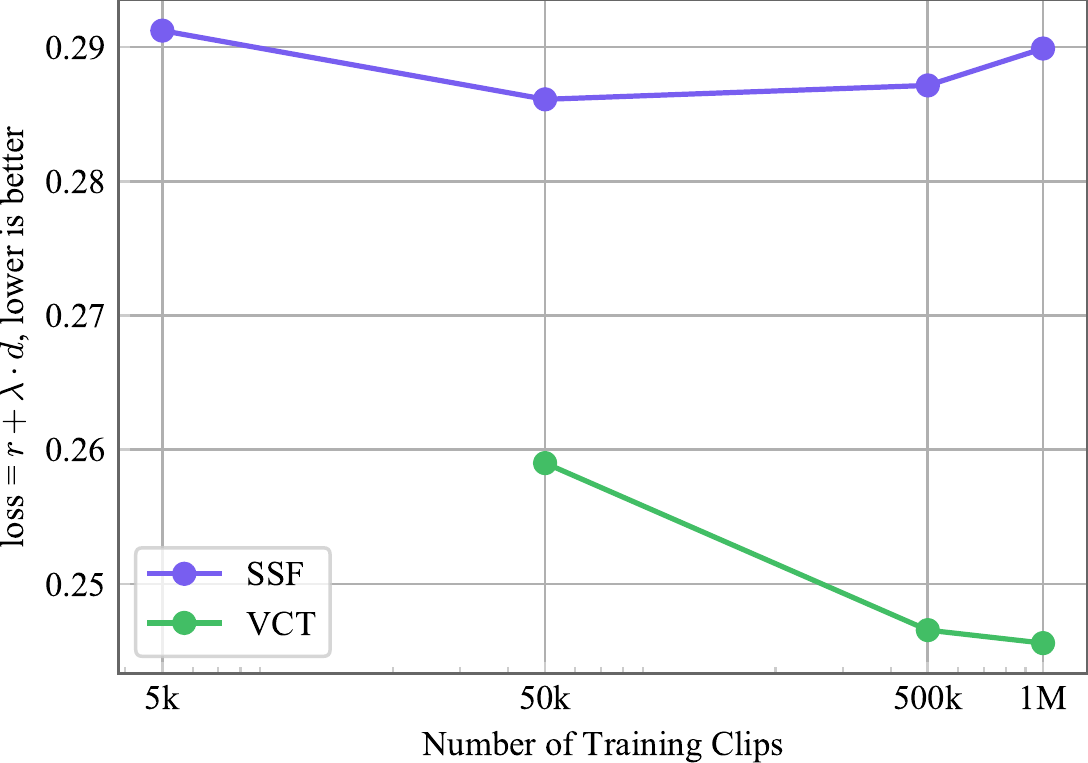}
    \caption{Effect of dataset size on evaluation loss on MCL-JCV. 
    In contrast to the transformer-based architecture,  SSF doesn't benefit from a significant increase in the amount of training data.
    \label{fig:num_ex}
    }
\end{figure}

In Fig.~\ref{fig:num_ex}, we show the effect
of dataset size on the loss on MCL-JCV for Ours (VCT) and the CNN based SSF~\cite{agustsson2020scale}.
We observe that VCT benefits from more training data, as has been observed when using transformers in other vision tasks~\cite{dosovitskiy2020image}. Note that 50k clips leads to VCT outperforming SSF.

\subsection{Transformers vs.\ CNNs}

\begin{figure}[h]
    \centering
    \includegraphics[width=0.5\linewidth]{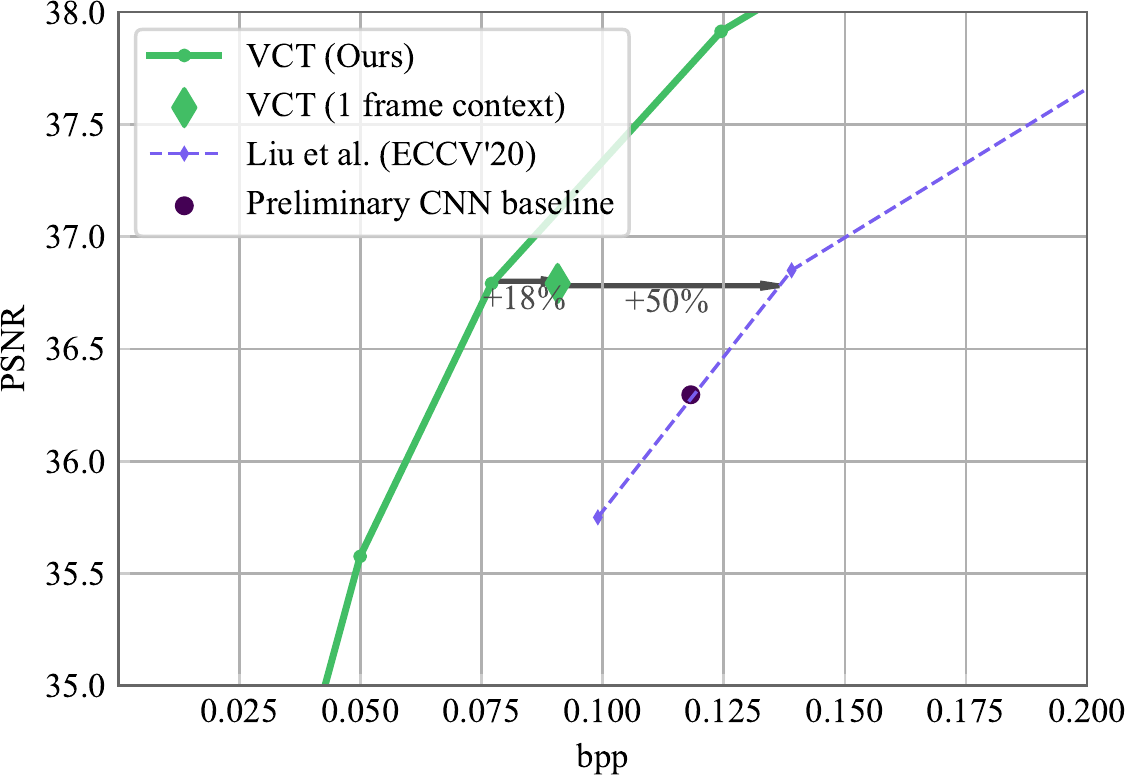}
    \caption{Comparing to Liu et al.~\cite{liu2020conditional}
    \label{fig:liucomp}}
    
\end{figure}

In Fig.~\ref{fig:liucomp}, we compare VCT against the CNN based method by Liu et al.~\cite{liu2020conditional}, which studies a similar setting as VCT, but uses CNNs for the temporal entropy model.
We provide preliminary results of reproducing Liu et al.'s work using CNNs, trained on our data (purple dot, denoted ``Preliminary CNN baseline'').
We can see that the baseline obtains a similar rate-distortion performance as the work by Liu et al.
Thus, similar to SFF (see Sec.~\ref{sec:app:data} above), we see that the CNN based approaches do not benefit from additional training data.

The main remaining differences to~\cite{liu2020conditional} are: i) they use 1 frame of context (vs.\ VCT's 2), ii) they rely on CNNs instead of our transformer.
We thus plot the model from the ablation study in Table~\ref{tab:context}, where we only use 1 frame as context, showing that this makes bitrate worse (green cross in Fig.~\ref{fig:liucomp}, $+18\%$ bitrate increase). 
From this, we can conclude that the transformer is responsible for the bulk of the remaining gap, \ie, the bitrate increases around 50\% when going to a CNN.

\subsection{Full-sized rate-distortion plots} \label{sec:app:bigplots}

In the following, we show a larger version of Fig.~\ref{fig:rd_results} to aid readability.

\newpage

\begin{figure}
    \centering
    \includegraphics[width=\textwidth]{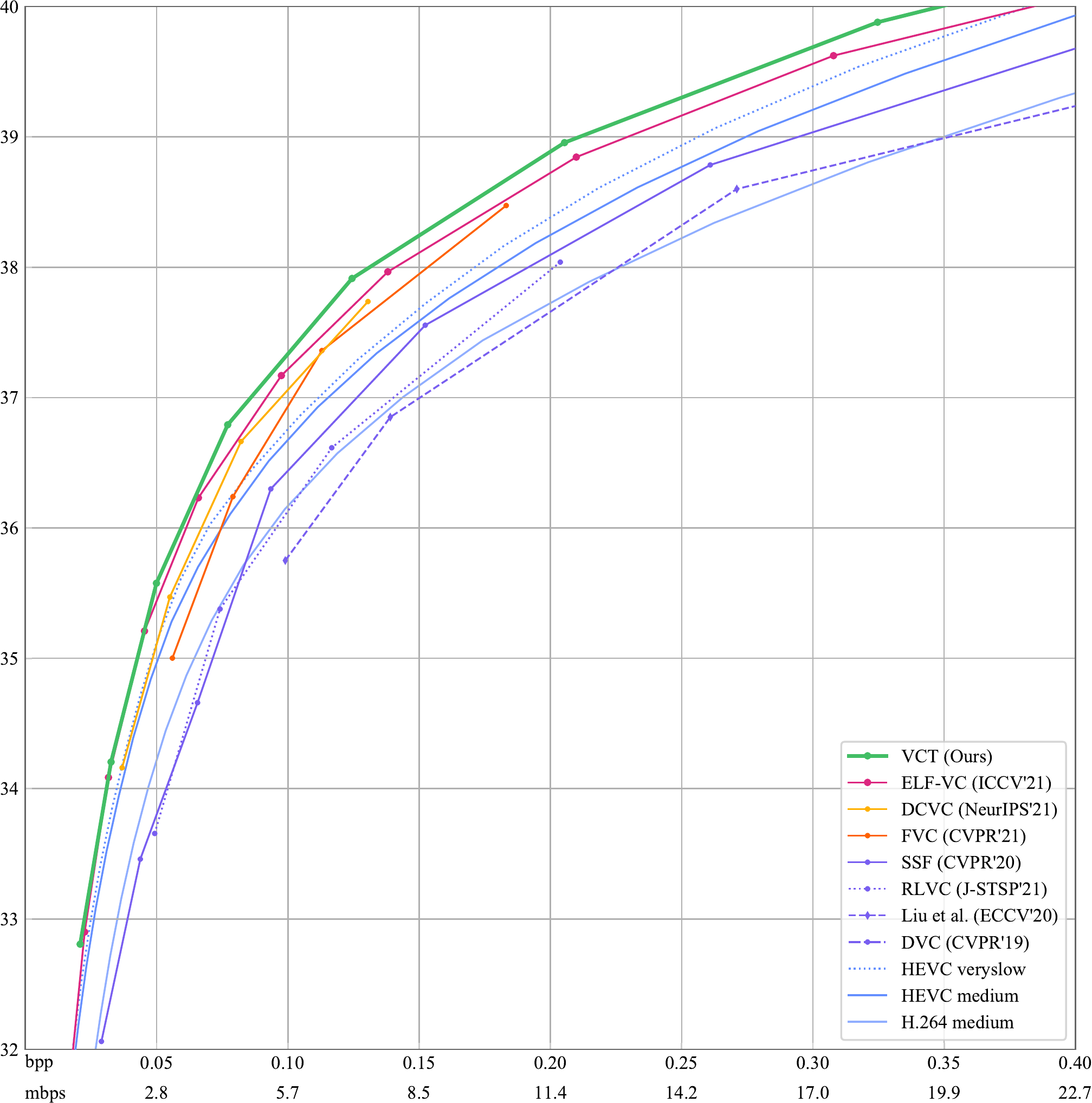}
    \caption{PSNR on MCL-JCV}
\end{figure}

\begin{figure}
    \centering
    \includegraphics[width=\textwidth]{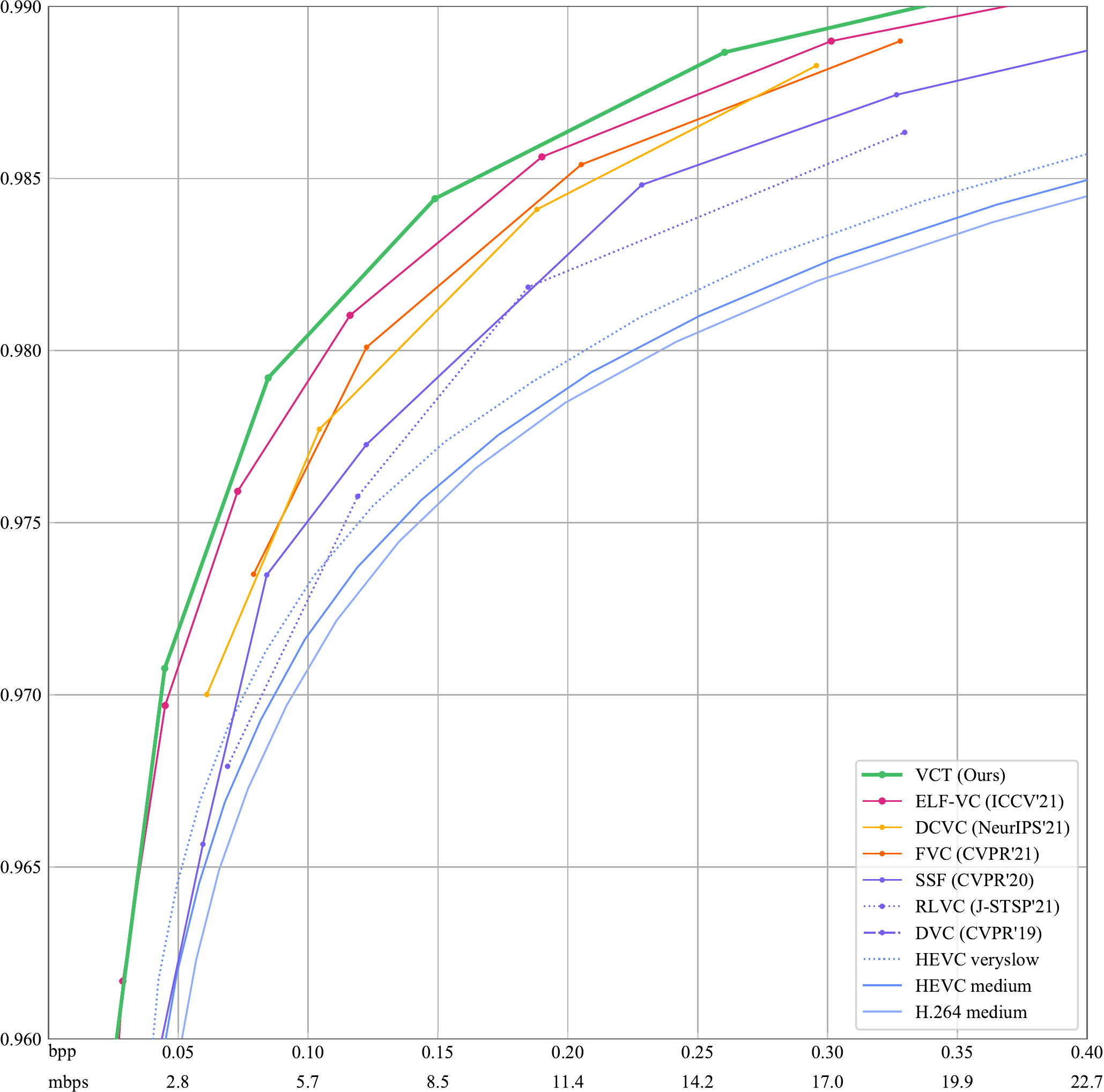}
    \caption{MS-SSIM on MCL-JCV}
\end{figure}

\begin{figure}
    \centering
    \includegraphics[width=\textwidth]{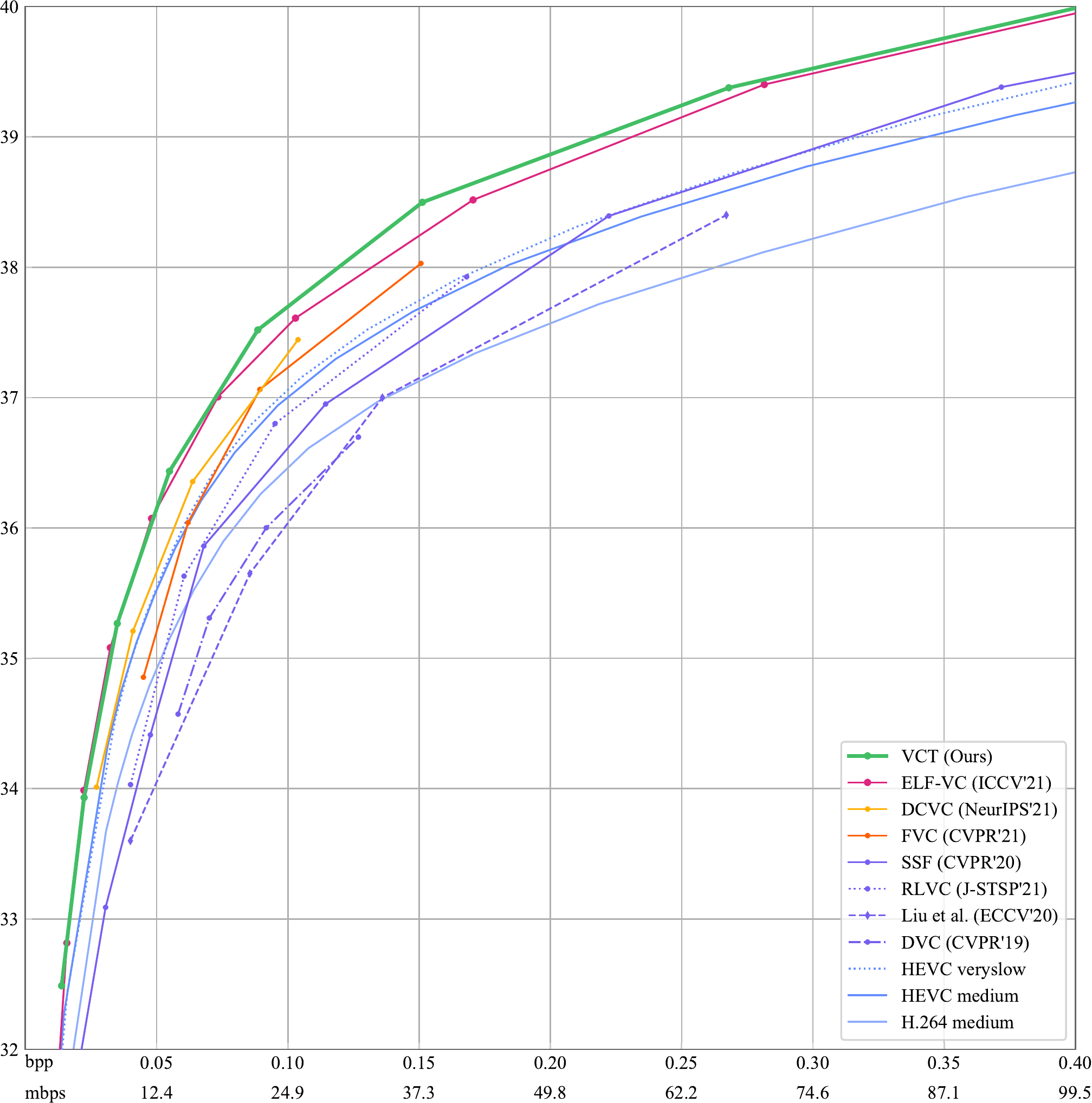}
    \caption{PSNR on UVG}
\end{figure}

\begin{figure}
    \centering
    \includegraphics[width=\textwidth]{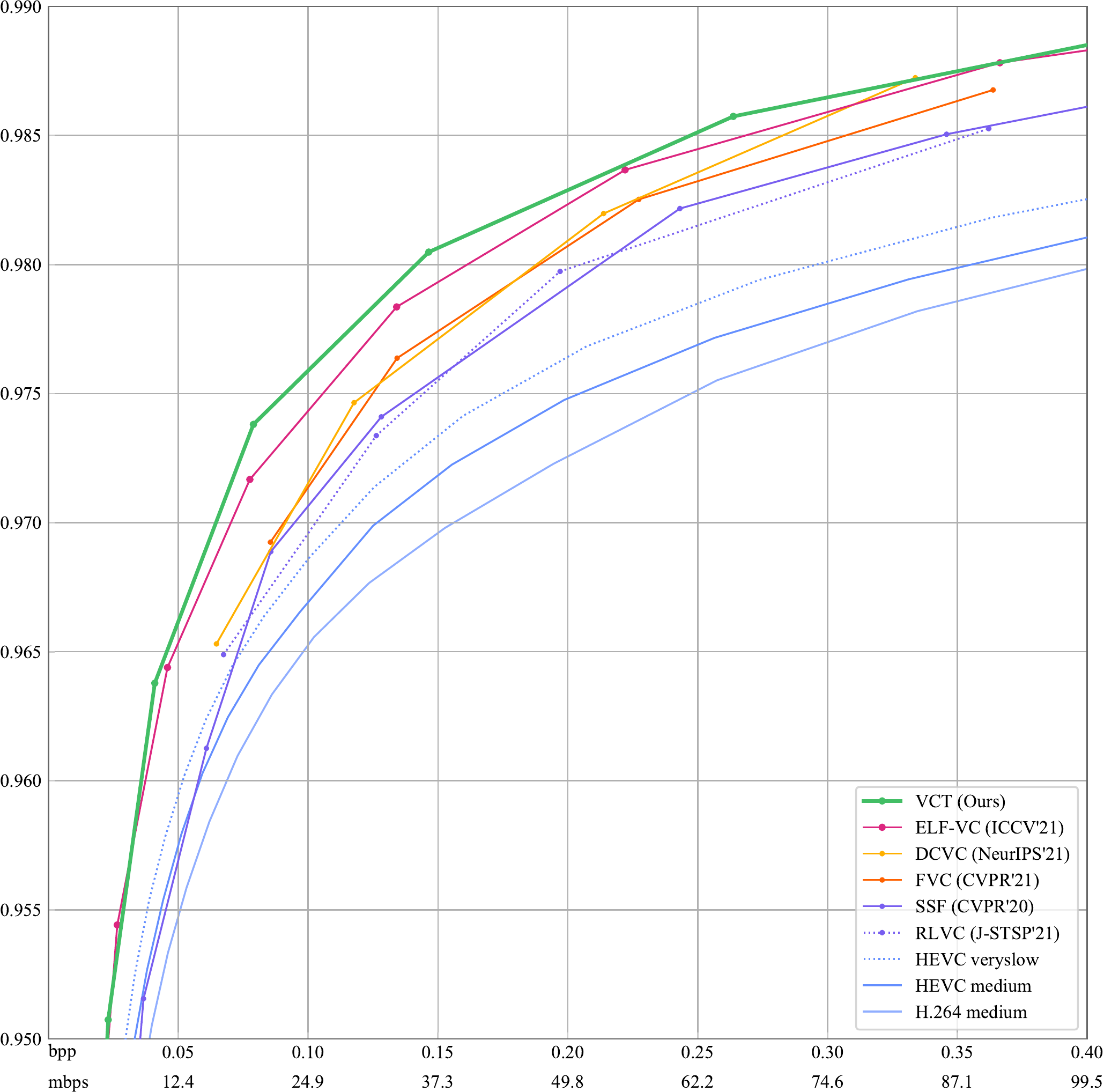}
    \caption{MS-SSIM on UVG}
\end{figure}

\end{document}